\documentclass[journal]{IEEEtran}
\usepackage{cite}
\ifCLASSINFOpdf
  \usepackage[pdftex]{graphicx}
\else
  \usepackage[dvips]{graphicx}
\fi
\usepackage[cmex10]{amsmath}
\DeclareMathOperator*{\argmin}{arg\,min}

\DeclareMathOperator*{\argmax}{arg\,max}

\usepackage{algorithmic}
\usepackage{algorithm}
\usepackage{color}

\usepackage{array}
\ifCLASSOPTIONcompsoc
  \usepackage[caption=false,font=normalsize,labelfont=sf,textfont=sf]{subfig}
\else
  \usepackage[caption=false,font=footnotesize]{subfig}
\fi
\usepackage{fixltx2e}
\usepackage{url}
\usepackage{hyperref}

\usepackage{amsfonts}
\usepackage{amsthm}

\newtheorem*{theorem*}{Theorem}

\newtheorem{lemma}{Lemma}

\newtheorem{fact}{Fact}

\newcommand{\specialcell}[2][c]{%
  \begin{tabular}[#1]{@{}c@{}}#2\end{tabular}}

\begin{document}
%
% paper title
% Titles are generally capitalized except for words such as a, an, and, as,
% at, but, by, for, in, nor, of, on, or, the, to and up, which are usually
% not capitalized unless they are the first or last word of the title.
% Linebreaks \\ can be used within to get better formatting as desired.
% Do not put math or special symbols in the title.
\title{Predicting Grades}
%
%
% author names and IEEE memberships
% note positions of commas and nonbreaking spaces ( ~ ) LaTeX will not break
% a structure at a ~ so this keeps an author's name from being broken across
% two lines.
% use \thanks{} to gain access to the first footnote area
% a separate \thanks must be used for each paragraph as LaTeX2e's \thanks
% was not built to handle multiple paragraphs
%

\author{Yannick~Meier,
        Jie~Xu,
        Onur~Atan,
        and~Mihaela~van~der~Schaar~\IEEEmembership{Fellow,~IEEE}% <-this % stops a space
\thanks{Copyright (c) 2015 IEEE. Personal use of this material is permitted. However, permission to use this material for any other purposes must be obtained from the IEEE by sending a request to \href{mailto:pubs-permissions@ieee.org}{pubs-permissions@ieee.org}.

Y. Meier, J. Xu, O. Atan and M. van der Schaar are with the Department of Electrical Engineering, University of California, Los Angeles, CA, 90095 USA. e-mail: (see \url{http://medianetlab.ee.ucla.edu/people.html}).

This research is supported by the US Air Force Office of Scientific Research under the DDDAS Program.}% <-this % stops a space
}

\maketitle

% As a general rule, do not put math, special symbols or citations
% in the abstract or keywords.
\begin{abstract}
To increase efficacy in traditional classroom courses as well as in Massive Open Online Courses (MOOCs), automated systems supporting the instructor are needed. One important problem is to automatically detect students that are going to do poorly in a course early enough to be able to take remedial actions. Existing grade prediction systems focus on maximizing the accuracy of the prediction while overseeing the importance of issuing timely and personalized predictions. This paper proposes an algorithm that predicts the final grade of each student in a class. It issues a prediction for each student individually, when the expected accuracy of the prediction is sufficient. The algorithm learns online what is the optimal prediction and time to issue a prediction based on past history of students' performance in a course. We derive a confidence estimate for the prediction accuracy and demonstrate the performance of our algorithm on a dataset obtained based on the performance of approximately 700 UCLA undergraduate students who have taken an introductory digital signal processing over the past 7 years. We demonstrate that for 85\% of the students we can predict with 76\% accuracy whether they are going do well or poorly in the class after the 4\textsuperscript{th} course week. Using data obtained from a pilot course, our methodology suggests that it is effective to perform early in-class assessments such as quizzes, which result in timely performance prediction for each student, thereby enabling timely interventions by the instructor (at the student or class level) when necessary.
\end{abstract}

% Note that keywords are not normally used for peerreview papers.

% For peer review papers, you can put extra information on the cover
% page as needed:
% \ifCLASSOPTIONpeerreview
% \begin{center} \bfseries EDICS Category: 3-BBND \end{center}
% \fi
%
% For peerreview papers, this IEEEtran command inserts a page break and
% creates the second title. It will be ignored for other modes.
\IEEEpeerreviewmaketitle

\section{Introduction}
% The very first letter is a 2 line initial drop letter followed
% by the rest of the first word in caps.
% 
% form to use if the first word consists of a single letter:
% \IEEEPARstart{A}{demo} file is ....
% 
% form to use if you need the single drop letter followed by
% normal text (unknown if ever used by IEEE):
% \IEEEPARstart{A}{}demo file is ....
% 
% Some journals put the first two words in caps:
% \IEEEPARstart{T}{his demo} file is ....
% 
% Here we have the typical use of a "T" for an initial drop letter
% and "HIS" in caps to complete the first word.
\IEEEPARstart{E}{ducation} is in a transformation phase; knowledge is increasingly becoming freely accessible to everyone (through Massive Open Online Courses, Wikipedia, etc.) and is developed by a large number of contributors rather than by a single author \cite{baraniuk2015}. Furthermore, new technology allows for personalized education enabling students to learn more efficiently and giving teachers the tools to support each student individually if needed, even if the class is large \cite{OpenStaxCollege}.
% You must have at least 2 lines in the paragraph with the drop letter
% (should never be an issue)

Grades are supposed to summarize in a single number or letter how well a student was able to understand and apply the knowledge conveyed in a course. Thus it is crucial for students to obtain the necessary support to pass and do well in a class. However, with large class sizes at universities and even larger class sizes in Massive Open Online Courses (MOOCs), which have undergone a rapid development in the past few years, it has become impossible for the instructor and teaching assistants to keep track of the performance of each student individually. This can lead to students failing in a class who could have passed if appropriate remedial actions had been taken early enough or excellent students not receiving the necessary promotion to benefit maximally from the course. Remedial or promotional actions could consist of additional online study material presented to the student in a personalized and/or automated manner \cite{tekinetutor}. Hence, in both offline and online education, it is of great importance to develop automated personalized systems that predict the performance of a student in a course before the course is over and as soon as possible. While in online teaching systems a variety of data about a student such as responses to quizzes, activity in the forum and study time can be collected, the available data in a practical offline setting are limited to scores in early performance assessments such as homework assignments, quizzes and midterm exams. 

In this paper we focus on predicting grades in traditional classroom-teaching where only the scores of students from past performance assessments are available. However, we believe that our methods can also be applied for online courses such as MOOCs. We design a grade prediction algorithm that finds for each student the best time to predict his/her grade such that, based on this prediction, a timely intervention can be made if necessary. Note that we analyze data from a digital signal processing course where no interventions were made; hence, we do not study the impact of inventions and consider only a single grade prediction for each student. However, our algorithm can be easily extended to multiple predictions per student.

A timely prediction exclusively based on the limited data from the course itself is challenging for various reasons. First, since at the beginning most students are motivated, the score of students in early performance assessments (e.g. homework assignments) might have little correlation with their score in later performance assessments, in-class exams and the overall score. Second, even if the same material is covered in each year of the course, the assignments and exams change every year. Therefore, the informativeness of particular assignments with regard to predicting the final grade may change over the years. Third, the predictability of students having a variety of different backgrounds is very diverse. For some students an accurate prediction can be made very early based on the first few performance assessments. If for example a student shows an excellent performance in the first three homework assignments and in the midterm exam, it is highly likely that he/she will pass the class. For other students it might take more time to make an equally accurate prediction. If a student for example performs below average but not terribly at the beginning, it is risky to predict whether he/she is going to pass or fail and, therefore, to decide whether or not to intervene. This third challenge illustrates the necessity to make the prediction for each student individually and not for all at the same time.

The main contributions of this paper can be summarized as follows.
\begin{enumerate}
	\item We propose an algorithm that makes a personalized and timely prediction of the grade of each student in a class. The algorithm can both be used in regression settings, where the overall score is predicted, and in classification settings, where the students are classified into two (e.g. do well/poorly) or more categories.
	\item We accompany each prediction with a confidence estimate indicating the expected accuracy of the prediction.
	\item We derive a bound for the probability that the prediction error is larger than a desired value $\epsilon$.
	\item We exclusively use the scores students achieve in early performance assessments such as homework assignments and midterm exams and do not use any other information such as age, gender or previous GPA. This makes our algorithm applicable in all practical traditional classroom and online teaching settings, where such information may not be available.
	\item Since the algorithm is learning from past years, the predictions become more accurate when more data from previous years become available.
	\item We demonstrate that the algorithm shows good robustness if different instructors have taught the course in past years.
	\item We analyze real data from an introductory digital signal processing course taught at UCLA over 7 years and use the data to experimentally demonstrate the performance of our algorithm compared to benchmark prediction methods. As benchmark algorithms we use well known algorithms such as linear/logistic regression and k-Nearest Neighbors, which are still a current research topic \cite{marjanovic2014l_,marano2013nearest, mateos2010distributed}.
 \item Based on our simulations, we suggest a preferred way of designing courses that enables early prediction and early intervention. Using data from a pilot course, we demonstrate the advantages of the suggested design.
\end{enumerate}

The rest of the paper is organized as follows. Section \ref{sec:related_work} discusses related work in the field of grade and GPA prediction in education. In Section \ref{sec:formalism_algorithm} we introduce notation, define data structures, formalize the problem and present the grade prediction algorithm. We analyze the data, describe benchmark methods and present simulation results including our and benchmark algorithms in Section \ref{sec:experiments}. Finally, we draw conclusions in Section \ref{sec:conclusion}.

\section{Related Work}
\label{sec:related_work}

Various studies have investigated the value of standardized tests \cite{kuncel2007standardized, cohn2004determinants, julian2005validity} admissions exams \cite{gallagher2001using} and GPA in previous programs \cite{cohn2004determinants} in predicting the academic success of students in undergraduate or graduate schools. They agree on a positive correlation between these predictors and success measures such as GPA or degree completion. Besides standardized tests, the relevancy of other variables for predictions of a student's GPA have been investigated, usually resulting in the conclusion that GPA from prior education and past grades in certain subjects (e.g. math, chemistry) \cite{gorr1994comparative,nghe2007comparative} have a strongly positive correlation as well. Reference \cite{gorr1994comparative} observes that simple linear and more complex nonlinear (e.g. artificial neural network) models frequently lead to similar prediction accuracies and concludes that there is either no complex nonlinear pattern to be found in the underlying data or the pattern cannot be recognized by their approach. Our simulations support the statement that simple linear models show a similar accuracy in grade predictions as more complex methods.

Reference \cite{goldman1976college} argues that the accuracy of GPA predictions frequently is mediocre due to different grading standards used in different classes and shows a higher validity for grade predictions in single classes. Consequently, many works focus on identifying relationships between a student's grade in a particular class and variables related to the student \cite{huang2013predicting, osmanbegovic2012data, baradwaj2012mining, yadav2012data, ahmed2014data, cortez2008using,werth1986predicting,turner1997factors,wang2002predictors,kotsiantis2004predicting}. Relevant factors were found to include the student's prior GPA \cite{huang2013predicting
, osmanbegovic2012data, cortez2008using, werth1986predicting, turner1997factors, kotsiantis2004predicting}, performance in related courses \cite{werth1986predicting, turner1997factors, kotsiantis2004predicting}, previous semester marks \cite{yadav2012data}, performance in entrance exams \cite{osmanbegovic2012data}, performance in early assignments of the class \cite{turner1997factors, kotsiantis2004predicting}, class attendance \cite{cortez2008using}, self-efficacy \cite{wang2002predictors} and whether the student is repeating the class \cite{turner1997factors}.

\begin{table*}[!t]
	\caption{Comparison With Related Work}
	\centering
		\begin{tabular}{ l | c | c | c | c | c | c}
														& \cite{werth1986predicting,wang2002predictors} & \cite{ahmed2014data, baradwaj2012mining, osmanbegovic2012data, yadav2012data, kotsiantis2004predicting} & \cite{huang2013predicting, cortez2008using,turner1997factors} &\cite{brintonmooc}	& \cite{romero2013predicting, lopez2012classification, calvo2006predicting, garcia2012promising, romero2008data, minaei2003predicting}& Our Work	\\
			\hline
			Goal of Paper												& \specialcell{Find Relevant\\Features} & \specialcell{Predict Course\\Grade} & \specialcell{Predict Course\\Grade} & \specialcell{Predict Accuracy\\of Answer} & \specialcell{Predict Course\\Grade}	& \specialcell{Predict Course\\Grade}	\\
			\hline
			Features											& Other	& Course \& Other	&	Course \& Other	& From Course				& From Course		& From Course \\
			\hline
			Learning from Past Years 			& n/a	& No							&	No							& No								& No						& Yes				\\
			\hline
			Accuracy-Timeliness Trade-Off	& n/a	& No					 &	No							& No								& No						& Yes	\\
			\hline
			Regression / Classification		& n/a	&	Classification	& Both					& Classification		& Classification & Both \\
		\end{tabular}
	\label{tab:ComparisonWithRelatedWork}
\end{table*}

A limitation of the algorithms in the previously discussed papers is that they are difficult to apply in many education scenarios. Frequently, variables related to the student such as performance in related classes, GPA or self-efficacy are not available to the instructor because the data has not been collected or is not accessible due to privacy reasons. However, the instructor always has access to data he/she collects from his/her own course, such as the performance of each student in early homework assignments or midterm exams. This paper, therefore, focuses on predicting the final grade based on this easily accessible data, which is collected anyway by the instructor.

Other works \cite{brintonmooc, romero2013predicting, lopez2012classification, garcia2012promising, romero2008data, calvo2006predicting, minaei2003predicting}, which also exclusively use data from the course itself, differ significantly from this paper in several aspects. First, they rely on logged data in online education or Massive Open Online Course (MOOC) systems such as information about video-watching behavior, time spent on specific questions or forum activity. In contrast, our results are applicable to both online and offline courses, which include some kind of graded assignments or related feedback from the students during the course. Second, in order for the instructor to be able to take corrective actions it is of great importance to predict with a certain confidence the performance of students as early as possible. While our algorithm takes this into account by deciding for each student individually the best time to make the prediction using a confidence measure, related works do not provide a metric indicating the optimal time to predict. Third, while related works need training data from the course whose grades they want to predict, we show that we can use training data from past year classes of the same course. Finally, in contrast to algorithms from related work, which are only shown to be applicable to classification settings (e.g. pass/fail or letter grade), our algorithm can be used both in regression and classification settings.

To make the predictions, related works use various data mining models such as regression models \cite{huang2013predicting, lopez2012classification}, decision trees \cite{ahmed2014data, romero2013predicting, osmanbegovic2012data, yadav2012data, lopez2012classification, baradwaj2012mining, minaei2003predicting}, support vector machines \cite{romero2013predicting, huang2013predicting, kotsiantis2004predicting, brintonmooc}, neural networks \cite{osmanbegovic2012data, calvo2006predicting, romero2008data}, Bayesian classifiers \cite{romero2013predicting, osmanbegovic2012data}, clustering \cite{lopez2012classification} and nearest neighbor techniques \cite{kotsiantis2004predicting, brintonmooc, romero2008data, minaei2003predicting }.

Table \ref{tab:ComparisonWithRelatedWork} summarizes the comparison between our paper and related work investigating and predicting student performance in a course.

\section{Formalism, Algorithm and Analysis}
\label{sec:formalism_algorithm}

In this section we mathematically formalize the problem and propose an algorithm that predicts the final score or a classification according to the final grade of a student with a given confidence.

\subsection{Definitions and System Description}

\begin{figure}[!t]
	\centering
		\includegraphics[width=3.4in]{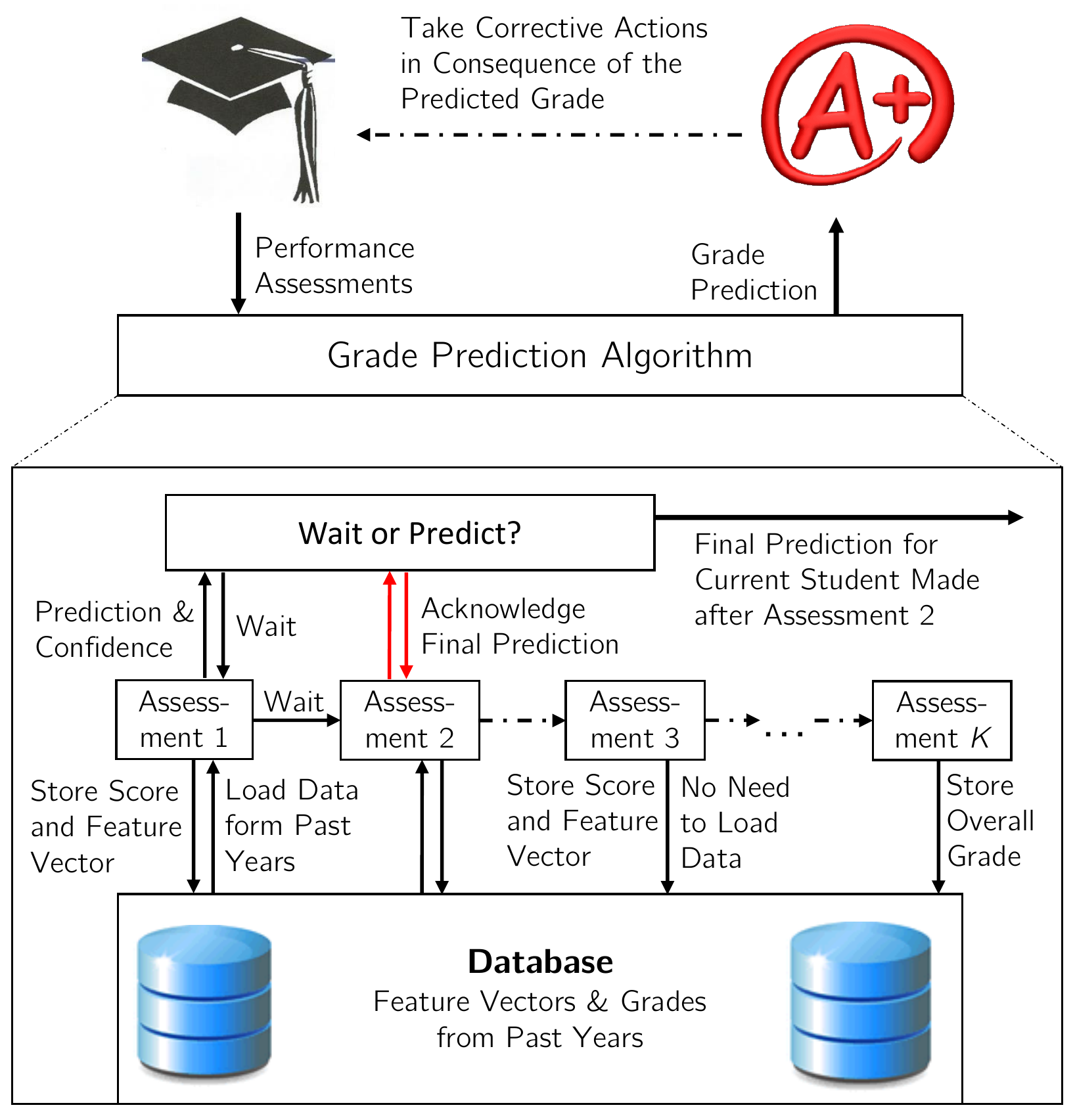}
	\caption{System diagram for a single student.}
	\label{fig:System_diagram}
\end{figure}

Consider a course which is taught for several years with only slight modifications. Students attending the course have to complete performance assessments such as graded homework assignments, course projects and in-class exams and quizzes throughout the entire course.\footnote{The performance assessments are usually graded by teaching assistants, by the instructor or even by other students through peer review \cite{xiao2014incentive}.} Our goal is to predict with a certain confidence the overall performance of a student before all performance assessments have been taken. See Fig. \ref{fig:System_diagram} for a depiction of the system.

We consider a discrete time model with $y=1,2,\ldots,Y$ and $k=1,2,\ldots,K$ where $y$ denotes the year in which the course is taught and $k$ the point in time in year $y$ after the $k$th performance assessment has been graded. $Y$ gives the total number of years during which the course is taught and $K$ is the total number of performance assessments of each year. For a given year $y$ we use index $i$ as a representation of $i$th student of the year and $I_y$ to denote the total number of students attending in year $y$. Except for the rare case that a student retakes the course, the students in each year are different. Let $a_{i,y,k}\in[0,1]$ denote the normalized score or grade of student $i$ in performance assessment $k$ of year $y$.

The feature vector of $y$th year student $i$ after having taken performance assessment $k$ is given by $\mathbf{x}_{i,y,k} =\left( a_{i,y,1},\ldots,a_{i,y,k} \right)$. The normalized overall score $z_{i,y}\in[0,1]$ of $y$th year student $i$ is the weighted sum of all performance assessments
\begin{equation}
z_{i,y} = \sum_{k=1}^K w_k a_{i,y,k}
\label{eq:exact_overall_score}
\end{equation}
where the $w_k$ denote the weight of performance assessment $k$ so that $\sum_{k=1}^K w_k =1$. The weights are set by the instructor and we assume that in each year the number, sequence and weight of performance assessments is the same. This assumption is reasonable since the content of a course usually does not change drastically over the years and frequently the same course material (e.g. course book) is used.\footnote{This assumption is made for simplicity. As we discuss in section \ref{subsec:implementation_alg} and show in Fig. \ref{fig:Cross_Instructor_Prediction} we can apply our algorithm to settings where different instructors using a different number and sequence of performance assessments and using different weights for each performance assessment teach the course. \label{ftn:assumption_about_courses}} This is especially true in an introductory course such as the one we investigate in Section \ref{sec:experiments}. The residual (overall score) $c_{i,y,k}$ of $y$th year student $i$ after performance assessment $k$ is defined as 
\begin{equation}
c_{i,y,k}= \begin{cases}
								\sum_{l=k+1}^K {w_l a_{i,y,l}}  & k\in \{1,\ldots,K-1\}	\\
								0																& k=K
						\end{cases}
\label{eq:def_residual}
\end{equation}
Using this definition we can write the overall score of $y$th year student $i$ as
\begin{equation}
z_{i,y}=c_{i,y,k} + \sum_{l=1}^k {w_l a_{i,y,l}}.
\label{eq:overall_residual}
\end{equation}
Note that after having taken the performance assessment $k$, the instructor has access to all the scores up to assignment $k$ but the residual scores $c_{i,y,k}$ need to be estimated. We denote the estimate of the residual score for $y$th year student $i$ at time $k$ by $\hat{c}_{i,y,k}$ and the corresponding estimate of the overall score by $\hat{z}_{i,y,k}$. In binary classification settings, where the goal is to predict whether a student achieves a letter grade above or below a certain threshold, we denote the class of $y$th year student $i$ by $b_{i,y} \in \left\{0,1\right\}$.

For each student $i$ we store the set of feature vectors $\mathbf{X}_{i,y}=\left\{\mathbf{x}_{i,y,k} | k\in \left\{1,\ldots,K \right\} \right\}$, the set of residuals $\mathbf{C}_{i,y}=\left\{c_{i,y,1},\ldots,c_{i,y,K-1}\right\}$ and the student's overall score $z_{i,y}$. All feature vectors from all students of year $y$ are given by $\mathbf{X}_y=\bigcup_{i=1}^{I_y} \mathbf{X}_{i,y}$ and $\mathbf{X}=\bigcup_{y=1}^Y \mathbf{X}_{y}$ denotes all feature vectors of all completed years. Similarly $\mathbf{C}=\bigcup_{y=1}^Y\bigcup_{i=1}^{I_y} \mathbf{C}_{i,y}$ and $\mathbf{Z}=\bigcup_{y=1}^Y\bigcup_{i=1}^{I_y} z_{i,y}$ denote all residuals and overall scores of all completed years. Let $\mathbf{X}^{k'}=\left\{\mathbf{x}_{i,y,k} |k=k',\forall i,y\right\}$ denote the set of feature vectors and $\mathbf{C}^{k'}=\left\{ c_{i,y,k}| k=k',\forall i,y\right\}$ denote the set of residuals saved after performance assessment $k'$.

\subsection{Problem Formulation}
Having introduced notations, definitions and data structures, we now formalize the grade prediction problem. We will investigate two different types of predictions. The objective of the first type, which we refer to as regression setting, is to accurately predict the overall score of each student individually in a timely manner. The second problem, referred to as classification setting, aims at making a binary prediction whether the student will do well or poorly or whether he/she will necessitate additional help or not. Again, the prediction is personalized and takes timeliness into account. For both types of predictions, the same algorithm can be used with only slight modifications, which we discuss in Section \ref{subsec:alg_classification}. We will also show that the binary prediction problem can easily be generalized to a classification into three or more classes.

Irrespective of the type of the prediction, the decision for a $y$th year student $i$ consists of two parts. First, we decide after which performance assessment $k_{i,y}^*$ to predict for the given student and second we determine his/her estimated overall score $\hat{z}_{i,y}$ or his/her estimated binary classification $\hat{b}_{i,y}$. At a point in time $k$ of year $y$ all scores including the overall scores of all students of past years $1,\ldots,y-1$ are known. Thus all feature vectors $\mathbf{x} \in \mathbf{X}$, residuals $c \in \mathbf{C}$ and overall scores $z \in \mathbf{Z}$ of all completed years are known. Furthermore, the scores $a_{i,y,1},\ldots,a_{i,y,k}$ of $y$th year student $i$ up to assessment $k$ are known as well and do not have to be estimated. However, to determine the overall score of the student we need to predict his/her residual score $c_{i,y,k}$ consisting of performance assessments $k+1,\ldots,K$ since they lie in the future and are unknown. At time $k$ we have to decide for each student of the current year whether this is the optimal time $k_{i,y}^*=k$ to predict or whether it is better to wait for the next performance assessment. If we decide to predict, we determine the optimal prediction of the overall score $\hat{z}_{i,y}=\hat{z}_{i,y,k_{i,y}^*}$. Both decisions are made based on the feature vector $\mathbf{x}_{i,y,k}$ of the given student and the feature vectors $\mathbf{x} \in \mathbf{X}^{k}$ and residuals $c \in \mathbf{C}^{k}$ of past students. To determine the optimal time to predict, we calculate a confidence $q_{i,y}(k)$ indicating the expected accuracy of the prediction for each student after each performance assessment. The prediction for a particular student is made as soon as the confidence exceeds a user-defined threshold $q_{i,y}(k) > q_{th}$. The problem of finding the optimal prediction time for $y$th year student $i$ is formalized as follows:
\begin{equation}
\begin{aligned}
& \underset{k}{\text{minimize}}
& & k \\
& \text{subject to}
& & q_{i,y}(k) > q_{th}
\end{aligned}
\label{eq:optimization_pred_time_student}
\end{equation}
The optimization problem results in the optimal prediction time $k_{i,y}^*$.

\subsection{Grade Prediction Algorithm, Regression Setting}
\label{subsec:alg_regression}

In this section we propose an algorithm that learns to predict a student's overall performance based on data from classes held in past years and based on the student's results in already graded performance assessments. We describe the algorithm for the regression setting and explain the changes needed to use the algorithm in the classification setting in Section \ref{subsec:alg_classification}.

Since at time $k$ we know the scores $a_{i,y,1},\ldots,a_{i,y,k}$ of the considered student from past performance assessments as well as the corresponding weights $w_{1},\ldots,w_{k}$, we only predict the residual $c_{i,y,k}$ and calculate the prediction of the overall score with (\ref{eq:overall_residual}). To make its prediction for the current residual of a student with feature vector $\mathbf{x}_{i,y,k}$, the algorithm finds all feature vectors from similar students of past years and their corresponding residuals $c_{i,y,k}$. We define the similarity of students through their feature vectors. Two feature vectors $\mathbf{x}_i, \mathbf{x}_j \in \mathbf{X}^k$ are similar if $\langle \mathbf{x}_i,\mathbf{x}_j \rangle_k \leq r$ where $\langle .,.\rangle_k$ is a distance metric defined on the feature space $\mathbf{X}^k$ and $r$ is a parameter. For two feature vectors $\mathbf{x}\in\mathbf{X}^{k_1}$ and $\mathbf{x'}\in\mathbf{X}^{k_2}$ from different feature spaces (i.e. $k_1 \neq k_2$) the distance metric is not defined since we only need to determine distances within a single feature space. Different feature spaces can have different definitions of the distance metric; we are going to define the distance metrics we use in Section \ref{subsec:implementation_alg}. We define a neighborhood $B\left( \mathbf{x}_c,r\right)$ with radius $r$ of feature vector $\mathbf{x}_c \in \mathbf{X}^k$ as all feature vectors $\mathbf{x}\in\mathbf{X}^k$ with $\langle \mathbf{x}_c,\mathbf{x} \rangle_k \leq r$.

Let $C^k$ denote the random variable representing the residual score after performance assessment $k$. $v^k\left(C^k|\mathbf{x}\right)$ denotes the probability distribution over the residual score for a student with feature vector $\mathbf{x}$ at time $k$ and $\mu^k(\mathbf{x})$ denotes the student's expected residual score. Let $p^k(\mathbf{x})$ denote the probability distribution of the students over the feature space $\mathbf{X}^k$. Intuitively $p^k(\mathbf{x})$ is the fraction of students with feature vector $\mathbf{x}$ at time $k$. Note that the distributions $v^k\left(C^k|\mathbf{x}\right)$ and $p^k(\mathbf{x})$ are not sampling distributions but unknown underlying distributions. We assume that the distributions do not change over the years.

We define the probability distribution of the students in a neighborhood $B\left( \mathbf{x}_c,r\right)$ with center $\mathbf{x}_c$ and radius $r$ as
\begin{equation*}
p^k_{\mathbf{x}_c,r}(\mathbf{x}):=\frac{p^k(\mathbf{x}) }{\int_{\mathbf{x}\in B(\mathbf{x}_c,r)} dp^k(\mathbf{x})} \mathbf{1}_{B(\mathbf{x}_c,r)}(\mathbf{x}),
\end{equation*}
where $\mathbf{1}$ is the indicator function. Intuitively $p^k_{\mathbf{x}_c,r}(\mathbf{x})$ is the fraction of students in neighborhood $B(\mathbf{x}_c,r)$ with feature vector $\mathbf{x}$. Let $C^k(B(\mathbf{x}_c,r))$ be the random variable representing the residual score of students in neighborhood $B(\mathbf{x}_c,r)$ after having taken performance assessment $k$. The distribution of $C^k(B(\mathbf{x}_c,r))$ is given by
\begin{equation*}
f^k_{\mathbf{x}_c,r}\left(C^k\right) := \int_{\mathbf{x}\in \mathbf{X}^k} v^k(C^k|\mathbf{x}) dp^k_{\mathbf{x}_c,r}(\mathbf{x})
\end{equation*}

We denote the true expected value of the residual scores after assignment $k$ of students in a particular neighborhood by $\mu^k\left(\boldsymbol{x}_{c},r\right) := \mathbb{E}(C^{k}\left( B\left(\mathbf{x}_{c},r\right)\right)) $. Note that
\begin{align*}
\mu^k\left(\boldsymbol{x}_{c},r\right) &=  \mathbb{E}_{\mathbf{x} \sim p^k_{\mathbf{x}_c,r}} \left[ \mathbb{E} \left[ C^k | \mathbf{x}\right] \right] = \mathbb{E}_{\mathbf{x} \sim p^k_{\mathbf{x}_c,r}} \left[ \mu^k \left( \mathbf{x}\right) \right] \\
& = \int_{\mathbf{x}\in \mathbf{X}^k}{\mu^k \left( \mathbf{x}\right) dp^k_{\mathbf{x}_c,r}}.
\end{align*}

 Our estimation of the true expected residual of students within a particular neighborhood $B(\boldsymbol{x}_{i,y,k},r)$ is given by 
\begin{align}
\hat{\mu}(C^{k}\left( B\left(\mathbf{x}_{i,y,k},r\right)\right)) = \frac{\sum\limits_{\mathbf{x} \in B\left(\mathbf{x}_{i,y,k},r\right)} c_{\mathbf{x},k}}{\left| B\left(\mathbf{x}_{i,y,k},r\right) \right|}
\label{eq:est_score}
\end{align}
where $c_{\mathbf{x},k}$ denotes the residual after time $k$ of the student with feature vector $\mathbf{x}$. For notational simplicity, we use $\hat{\mu}^k(\boldsymbol{x}_{i,y,k},r) := \hat{\mu}(C^{k}\left( B\left(\mathbf{x}_{i,y,k},r\right)\right))$ to denote the estimated expectation. In the following we are going to derive how confident we are in the estimation of the residual score based on a given neighborhood $B(\mathbf{x},r)$ and how we use this confidence $q\left(B(\mathbf{x},r)\right)$ to both select the optimal radius of the neighborhood and to decide when to predict.

Intuitively, if the feature vectors after performance assessment $k$ in a neighborhood $B(\mathbf{x},r)$ of $\mathbf{x}$ contain a lot of information about the residual $c_{\mathbf{x},k}$, past students with feature vectors in this neighborhood should have had similar residuals. Hence, the variance of the residuals $\text{Var}\left(C^k(B(\boldsymbol{x}_{i,y,k},r))\right)$ of the students in the neighborhood should be small. To mathematically support this intuition, we consider the residuals $c_{i,y,k}$ in a neighborhood $B(\mathbf{x},r)$ of feature vector $\mathbf{x}$ with distribution $f^k_{\mathbf{x}_c,r}\left(C^k\right)$. For any confidence interval $\epsilon$ the probability that the absolute difference between the unknown residual $c_{\mathbf{x},k}$ of the student with feature vector $\mathbf{x}$ and the expected value of the residual distribution $\mu^k\left( \mathbf{x}, r \right)$ in his/her neighborhood is smaller than $\epsilon$ can be bounded by
\begin{equation}
P\left[ |C^k(B(\boldsymbol{x},r)) - \mu^k(\boldsymbol{x},r) | < \epsilon \right] > 1-\frac{Var\left(C^k(B(\boldsymbol{x},r))\right)}{\epsilon^2}.
\label{eq:chebyshev}
\end{equation}
This statement directly follows from Chebyshev's inequality.

We conclude that the lower the variance of the residual distribution in the neighborhood, the more confident we are that the true residual $c_{\mathbf{x},k}$ will be close to $\mu^k(\mathbf{x},r)$. Since both the expected value $\mu^k(\mathbf{x},r)$ and the variance $\text{Var}\left(C^k(B(\boldsymbol{x},r))\right)$ of the distribution are unknown, we estimate the two values through the sample mean from (\ref{eq:est_score}) and the sample variance $\widehat{Var}\left(C^k(B(\boldsymbol{x},r))\right)$ given by
\begin{equation}
\widehat{Var}\left(C^k(B(\boldsymbol{x},r))\right) = \frac{\sum_{\mathbf{x} \in B(\mathbf{x},r)}{\left( c_{\mathbf{x},k} - \hat{\mu}^k(\boldsymbol{x},r)  \right)^2}}{|B(\mathbf{x},r)|-1}.
\label{eq:sample_var}
\end{equation}
In the following we use $Var^k(\boldsymbol{x},r):=\text{Var}\left(C^k(B(\boldsymbol{x},r))\right)$ to denote the variance and $\widehat{Var}^k(\mathbf{x},r):=\widehat{Var}\left(C^k(B(\boldsymbol{x},r))\right)$ to denote the sample variance of the residual distribution in neighborhood $B(\mathbf{x},r)$. From the law of large number it follows that the sample mean and the sample variance converge to the true expected value and the true variance for $|B(\mathbf{x},r)| \rightarrow \infty$. We will provide a bound for the probability that the prediction error is larger than a given value in the theorem below. Given a desired confidence interval $\epsilon$, we define the confidence on the prediction of the residual as 
\begin{equation}
q\left(B(\mathbf{x},r)\right) = 1-\frac{\widehat{Var}^k\left(\boldsymbol{x},r\right)}{\epsilon^2}.
\label{eq:confidence_var}
\end{equation}

Using this confidence measure the radius of the optimal neighborhood after performance assessment $k$ is given by $r^*=\argmax_{r} q\left( B\left(\mathbf{x}_{i,y,k},r\right) \right) = \argmin_r \widehat{Var}^k\left(\boldsymbol{x},r\right)$. To estimate $r^*$ after each performance assessment $k$, our algorithm considers $M$ different neighborhoods $B(\mathbf{x}_{i,y,k},r_m),m=1,\ldots,M$ with user-defined radii $r_m$ and chooses the best neighborhood $\hat{m}_k(\mathbf{x}_{i,y,k})$ according to our confidence measure $\hat{m}_k(\mathbf{x}_{i,y,k})=\argmax_m q\left(B(\mathbf{x}_{i,y,k},r_m)\right)$. In the following we use $\hat{m}_k:=\hat{m}_k(\mathbf{x}_{i,y,k})$ to denote the best neighborhood. Let
\begin{equation}
	\hat{c}_{i,y,k}:=\hat{\mu}^k \left(\mathbf{x}_{i,y,k},r_{\hat{m}_k}\right)
\label{eq:notation_best_residual}
\end{equation}
denote the estimated residual of the best neighborhood at time $k$ and $\hat{z}_{i,y,k}$ denotes the corresponding estimated overall score
\begin{equation}
\hat{z}_{i,y,k}=\hat{c}_{i,y,k} + \sum_{l=1}^k {w_l a_{i,y,l}}.
\label{eq:est_overall_score}
\end{equation}
  If the confidence bound for the best neighborhood $q_{i,y}(k)=q\left( B\left(\mathbf{x}_{i,y,k},r_{\hat{m}_k}\right) \right)$ is above a given threshold $q_{i,y}(k) \geq q_{th}$, the algorithm returns the final prediction of the overall score $\hat{z}_{i,y}=\hat{z}_{i,y,k}$ for the considered student.
	
\begin{figure}[!t]
	\centering
		\includegraphics[width=3.3in]{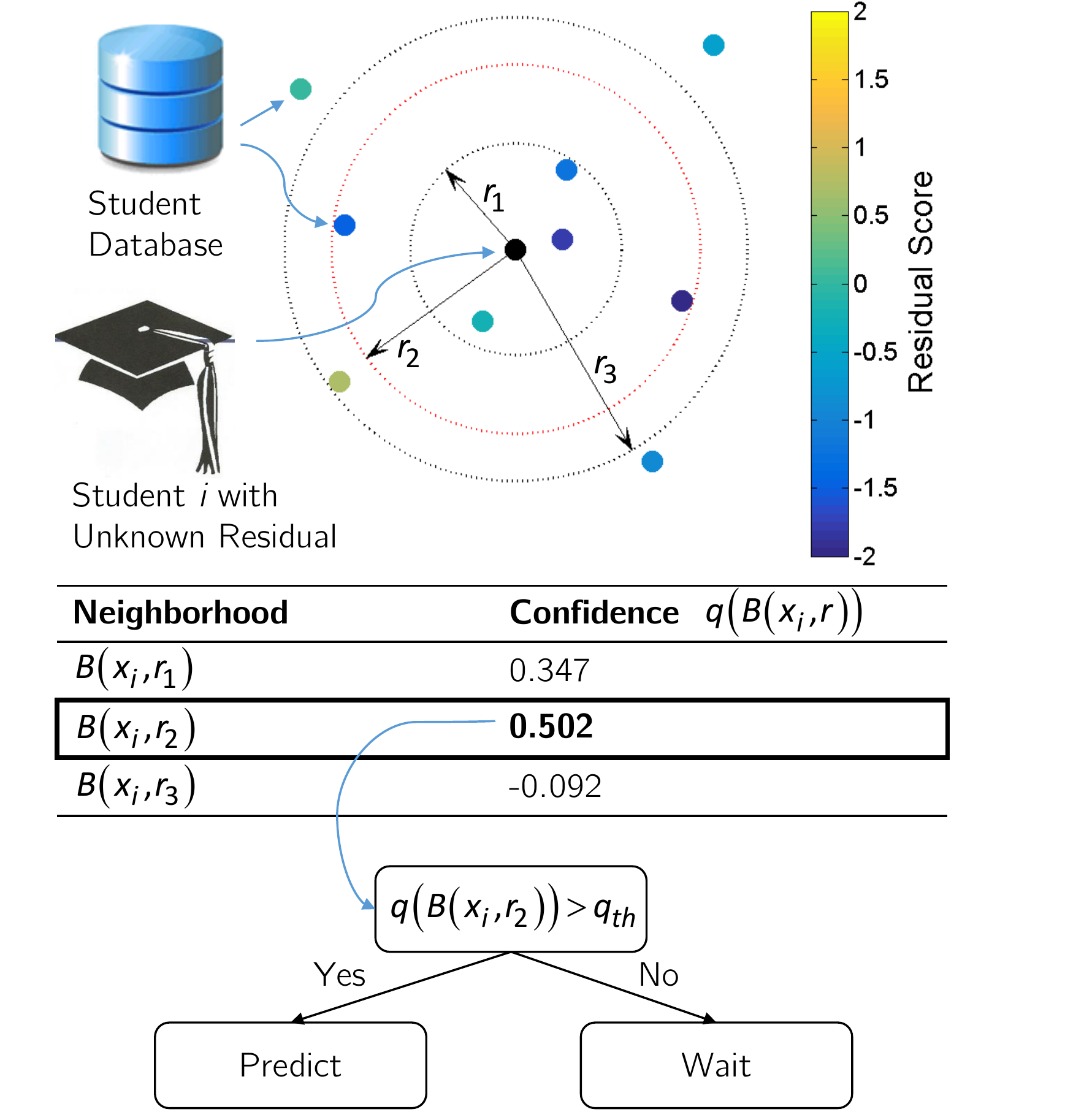}
	\caption{Illustration of the neighborhood selection process.}
	\label{fig:Illustrative_Figure_Balls}
\end{figure}

If the confidence is below the threshold, we wait for the next performance assessment and start the next iteration. Fig. \ref{fig:Illustrative_Figure_Balls} illustrates the neighborhood selection process.
Algorithm \ref{alg:cagp_variance} provides a formal description of the grade prediction algorithm in pseudocode.

\begin{algorithm}[!t]
\caption{Grade Prediction Algorithm, Regression Setting}
\label{alg:cagp_variance}
\begin{algorithmic}[1]
\REQUIRE All $\mathbf{x}$ and $z$ from past years, $q_{th}$, number $M$ and radii $r_1,\ldots,r_M$ of neighborhoods 
\ENSURE Predictions $\hat{z}$ for the overall scores of the students
\FORALL{years $y$}
	\FORALL{performance assessments $k$}
		\FORALL{current-year students $i$ for whom the final prediction has not been made}
			\IF{$k=K$}
				\STATE Calculate $z_{i,y}$ according to (\ref{eq:exact_overall_score})
				\STATE Return $z_{i,y}$ as final prediction for student $i$
			\ENDIF
			\STATE Create $M$ neighborhoods with radii $r_1,\ldots,r_M$
				\FORALL{neighborhoods $m$}
					\STATE Estimate residual $\hat{c}\left( B\left(\mathbf{x}_{i,y,k},r_m\right)\right)$ with (\ref{eq:est_score})
					\STATE Compute $\widehat{Var}\left(\mathbf{x},r_m\right)$ with (\ref{eq:sample_var})
					\STATE Compute $q\left(B(\mathbf{x}_{i,y,k},r_m)\right)$ with (\ref{eq:confidence_var})
				\ENDFOR
				\STATE Find $\hat{m}_k=\argmax_{m}q\left( B\left(\mathbf{x}_{i,y,k},r_m\right) \right)$ \label{line:find_neighborhood}
				\IF{$q\left( B\left(\mathbf{x}_{i,y,k},r_{\hat{m}_k}\right) \right) \geq q_{th}$} \label{line:confidence_decision} \label{line:prediction_decision}
					\STATE Compute $\hat{z}_{i,y}$ with (\ref{eq:overall_residual}) \label{line:return_minus_1}
					\STATE Return $\hat{z}_{i,y}$ as final prediction for student $i$ \label{line:return}
				\ENDIF
				\STATE Add $\mathbf{x}_{i,y,k}$ and $a_{i,y,k}$ to database
		\ENDFOR
	\ENDFOR
	\STATE Calculate all $c_{i,y,k}$ of year $y$ according to (\ref{eq:def_residual})
	\STATE Add all $c_{i,y,k}$ to database
\ENDFOR
\end{algorithmic}
\end{algorithm}

To conclude the discussion of the grade prediction algorithm in the regression setting, we derive a bound for the probability that the prediction error is larger than a value $\epsilon$. Before we state the theorem, we introduce some further notations. Let $m_k^*(\mathbf{x})$ denote the index of the neighborhood with the smallest variance of residuals for the student with feature vector $\mathbf{x}$ at time $k$
\begin{equation}
m_k^*(\mathbf{x})=\underset{1 \leq m \leq M}{\argmin} \  Var^k(\mathbf{x},r_m).
\label{eq:def_hood_smallest_var}
\end{equation}
Note that $m_k^*(\mathbf{x})$ is not necessarily equal to $\hat{m}_k(\mathbf{x})$, the index of the neighborhood with the highest confidence chosen by our algorithm, since the confidence defined in (\ref{eq:confidence_var}) is calculated with the known sample variance of residuals $\widehat{Var}(\mathbf{x},r)$ and not with the unknown true variance $Var^k(\mathbf{x},r)$ used in (\ref{eq:def_hood_smallest_var}).

Similarly $m_{k,2}^*(\mathbf{x})$ denotes the index of the neighborhood with the second highest confidence.
\begin{equation*}
m_{k,2}^*(\mathbf{x})=\underset{1 \leq m \leq M,m \neq m_k^*(\mathbf{x})}{\argmin} \  Var^k(\mathbf{x},r_m).
\end{equation*}
Let $\Delta_k(\mathbf{x})$ denote the difference between the standard deviations of the residual distribution of neighborhoods $m_k^*(\mathbf{x})$ and $m_{k,2}^*(\mathbf{x})$
\begin{equation}
\Delta_k(\mathbf{x})=\sqrt{Var^k(\mathbf{x},r_{m_{k,2}^*})} - \sqrt{Var^k(\mathbf{x},r_{m_{k}^*})}.
\label{eq:Delta_definition}
\end{equation}

\begin{theorem*}
Without loss of generality we assume that all scores $a$ are normalized to the range $[0,1]$. Consider the prediction $\hat{z}_{i,y,k}$ of the overall score of $y$th year student $i$ with feature vector $\mathbf{x}$ made by algorithm \ref{alg:cagp_variance}. The probability that the absolute error the prediction exceeds $\epsilon$ is bounded by
\begin{align*}
& P \left[ \left| z_{i,y} - \hat{z}_{i,y,k} \right|  \geq \epsilon \right] \leq \frac{4 Var^k\left(\mathbf{x},r_{m_k^*(\mathbf{x})}\right)}{\epsilon^2} \\
& \quad + 2 \exp\left[-\epsilon^2 \underset{1 \leq m \leq M}{\min} \frac{\left|B\left(\mathbf{x},r_{m}\right)\right|}{2}\right] \\
& \quad + 2 M \exp\left[- \Delta_k(\mathbf{x})^2 \underset{1 \leq m \leq M}{\min} \frac{|B(\mathbf{x},r_m)|-1}{8}\right] \\
\end{align*}
\end{theorem*}
\begin{IEEEproof}
See Appendix. 
\end{IEEEproof}

This theorem illustrates two important aspects of algorithm \ref{alg:cagp_variance}. First, we see that for a given neighborhood the accuracy of our predictions increases with an increasing number of neighbors. Hence, our algorithm learns the best predictions online as the knowledge base is expanded after each year, when the feature vectors and results from the past-year students are added to the database. In Section \ref{subsubsec:results_regression_setting} we show that this learning can be experimentally illustrated with our data from the digital signal processing course taught at UCLA. Second, the term $Var^k(\mathbf{x},r_{m_k^*})/\epsilon^2$ shows that the prediction accuracy will be higher if the variance of the residuals in a neighborhood is small. With increasing time $k$ we expect this variance to decrease since we have more information about the students and we expect the students in a neighborhood to be more similar and achieve similar (residual) scores.

Note that it is possible to restrict the data kept in the knowledge base to recent years, which allows the algorithm to adapt faster to slowly changing students and to changes in the course.

\subsection{Grade Prediction Algorithm, Classification Setting}
\label{subsec:alg_classification}

In the binary classification setting we predict the overall score analogously to the regression setting and then determine the class by comparing the predicted overall score $\hat{z}_{i,y}$ with a threshold score $z_{th}$. To illustrate how we find $z_{th}$ let us assume that we want to predict whether a student does well (letter grades $\geq B-$) or does poorly (letter grades $\leq C+$). To determine $z_{th}$, we find the average $z_{avg,B-}$ of all students from past years who received a $B-$ and the average $z_{avg,C+}$ of all students from past years who achieved a $C+$. Subsequently, we define $z_{th}$ as $z_{th}=\left(z_{avg,B-}+z_{avg,C+}\right)/2$.
The predicted classification $\hat{b}_{i,y}$ of $y$th year student $i$ is then given by
\begin{equation}
\hat{b}_{i,y}= \begin{cases}
								0			& \hat{z}_{i,y} \geq z_{th}	\\
								1			& \hat{z}_{i,y} < z_{th}.
						\end{cases}
\label{eq:classification_prediction}
\end{equation}
We are more confident in the classification not only if the variance of the neighbor-scores is small, which is the metric we used for the confidence in the regression setting, but also if the distance $d\left(B(\mathbf{x},r)\right)=\left| \hat{z}\left(B(\mathbf{x},r)\right)-z_{th} \right|$ between the predicted score and the threshold score is large. Note that $\hat{z}\left(B(\mathbf{x},r)\right)$ is the estimate of the overall score based on neighborhood $B(\mathbf{x},r)$. Because of this intuition we use a modified confidence
\begin{equation}
q^{bin}\left(B(\mathbf{x},r)\right) = 1-e^{-d\left(B(\mathbf{x},r)\right)} \frac{\widehat{Var}\left(C^k(B(\boldsymbol{x},r))\right)}{\epsilon^2}
\label{eq:confidence_var_binary}
\end{equation}
to decide whether to make the final prediction in binary classification settings. Since $d\left(B(\mathbf{x},r)\right)$ should only influence whether the final prediction is made for a given neighborhood but not the neighborhood selection process, we still use the unmodified confidence from (\ref{eq:confidence_var}) to select the optimal neighborhood.

In summary, four changes have to be made to algorithm \ref{alg:cagp_variance} to make it applicable to binary classification settings. First, $z_{th}$ has to be determined/updated at the beginning of each new year. Second, we calculate $\hat{b}_{i,y}$ after line \ref{line:return_minus_1} according to (\ref{eq:classification_prediction}). Third, we return $\hat{b}_{i,y}$ at line \ref{line:return} instead of $\hat{z}_{i,y}$. Fourth, we use the modified confidence $q^{bin}$ according to (\ref{eq:confidence_var_binary}) in line \ref{line:prediction_decision} instead of the unmodified confidence $q$. We use the unmodified confidence from (\ref{eq:confidence_var}) in line \ref{line:find_neighborhood}.

The described binary classification algorithm can easily be generalized to a larger number of categories. In a classification with $L$ categories, we define $L-1$ threshold values $z_{th,1}<z_{th,2}<\ldots<z_{th,L-1}$ and determine in which of the $L$ score intervals $\left\{[0,z_{th,1}),[z_{th,1},z_{th,2}),\ldots,[z_{th,L-1},1]\right\}$ the predicted overall score $\hat{z}_{i,y}$ of a student lies. The index of the interval corresponds to the classification of the student. In this general classification setting, the modified confidence from (\ref{eq:confidence_var_binary}) can be used as well by defining $d$ as the distance of $\hat{z}_{i,y}$ to the nearest threshold value.

We discuss the performance of the proposed algorithm \ref{alg:cagp_variance} in both regression and classification settings in Section \ref{subsec:results}.

\subsection{Confidence-Learning Prediction Algorithm}

Besides the radii of the neighborhoods $r_i$, the only parameter to be chosen by the user in algorithm \ref{alg:cagp_variance} is the desired confidence threshold $q_{th}$. Since for an instructor it is more natural and practical to specify a desired prediction accuracy or error directly rather than the confidence threshold, we show in this section how to automatically learn the appropriate confidence threshold to achieve a certain prediction performance and what consequences this has on the average prediction time. We will discuss a possible way of choosing the radii $r_i$ of the neighborhoods in Section \ref{subsec:implementation_alg}.

Formally we define the problem as follows. Let $p(k,q_{th}) \in \left[0,1\right]$ denote the proportion of current year students for which the grade prediction algorithm working with confidence threshold $q_{th} \leq 1$ has predicted the overall score by time (performance assessment) $k \in \left[0,K\right]$. $p_{min}$ is the minimum percentage of current year students whose grade the user wants to predict with a specified accuracy. $E(k,q_{th})\geq 0$ denotes the average absolute prediction error up to time $k$ for the given confidence. $E_{max}$ is the maximum error the user is willing to tolerate. $k(p,q_{th})$ is the time necessary to predict the grade for proportion $p$ of all students of the class using confidence threshold $q_{th}$. Please note that since the variables $p$, $E$, $q_{th}$ and $k$ are dependent, we can only independently specify two of the four variables. If we for example specify to predict all ($p=1$) students with zero error ($E=0$), the algorithm will have to wait until the end of the course when the overall score is known ($k=K$) and will use maximum confidence ($q_{th}=1$). Without making any assumptions on the dependence of the variables of each other, multiple pairs $(k,q_{th})$ might lead to the same specified pair $(p,E)$.

 Our goal is, therefore, to learn from past data the $y$th year estimate of the minimal time $k_y$ and corresponding confidence threshold $q_{th,y}$ necessary to achieve the desired share $p_{min}$ of students predicted and the desired maximum average prediction error $E_{max}$. This is formally defined as:

\begin{equation}
\begin{aligned}
& \underset{q_{th}}{\text{minimize}}
& & k(p,q_{th}) \\
& \text{subject to}
& & p(k,q_{th}) \geq p_{min} \\
& & & E(k,q_{th}) \leq E_{max}
\end{aligned}
\label{eq:optimization_for_prediction_time}
\end{equation}
Note that while the goal of optimization problem (\ref{eq:optimization_pred_time_student}) is to find the minimum time to predict the overall score of a particular student with a desired confidence, this problem (\ref{eq:optimization_for_prediction_time}) aims at finding the minimum time by which the overall scores of a specific percentage of all students can be predicted with a desired maximum error.

At a given year $y$ we solve this optimization problem using a brute force approach using the all available data from years $1,\ldots,y$. For this purpose, we extract $k$, $p$ and $E$ from algorithm \ref{alg:timeliness} for a large number of different confidence thresholds $q_{th}$. We then select the confidence threshold $q_{th,y}$ which is optimal with respect to optimization problem (\ref{eq:optimization_for_prediction_time}) and determine the corresponding prediction time $k_y$. To make the grade predictions for year $y+1$ we use the learned confidence threshold $q_{th,y}$ as input to prediction algorithm \ref{alg:cagp_variance}. Since there are no training data available yet at year $y=1$, the algorithm uses a user-defined starting value $q_{th,0}$ for the grade predictions of the first year. Algorithm \ref{alg:timeliness} summarizes the learning algorithm in pseudocode.

\begin{algorithm}[!t]
\caption{Confidence-Learning Prediction Algorithm}
\label{alg:timeliness}
\begin{algorithmic}[1]
\REQUIRE $E_{max}$, $p_{min}$, $q_{th,0}$, all $\mathbf{x}$ and $z$ from past years, number $M$ and radii $r_1,\ldots,r_M$ of neighborhoods
\ENSURE Predictions $\hat{z}$, $k_y$ and $q_{th,y}$ for all years
\FORALL{years $y$}
		\IF{$y>1$}
				\STATE Find $k_{y-1}$ and $q_{th,y-1}$ according (\ref{eq:optimization_for_prediction_time}) by running algorithm \ref{alg:cagp_variance} with various $q_{th}$
				\STATE Return $k_{y-1}$ and $q_{th,y-1}$
		\ENDIF
		\STATE Use algorithm \ref{alg:cagp_variance} with $q_{th}=q_{th,y-1}$ to predict and return the grades of current year $y$ students
\ENDFOR
\end{algorithmic}
\end{algorithm}

\section{Experiments}
\label{sec:experiments}

In this section, we present the data, discuss details of the application of algorithm \ref{alg:cagp_variance} to our dataset, illustrate the functioning of the algorithm and evaluate its performance by comparing it against other prediction methods in both regression and binary classification settings. Due to space limitations, we will not show experimental results for classification settings with more than two categories.

\subsection{Data Analysis}
\label{subsec:dataAnalysis}

\begin{figure*}[!t]
\centering
\subfloat[Grade distribution]{\includegraphics[width=2.3in]{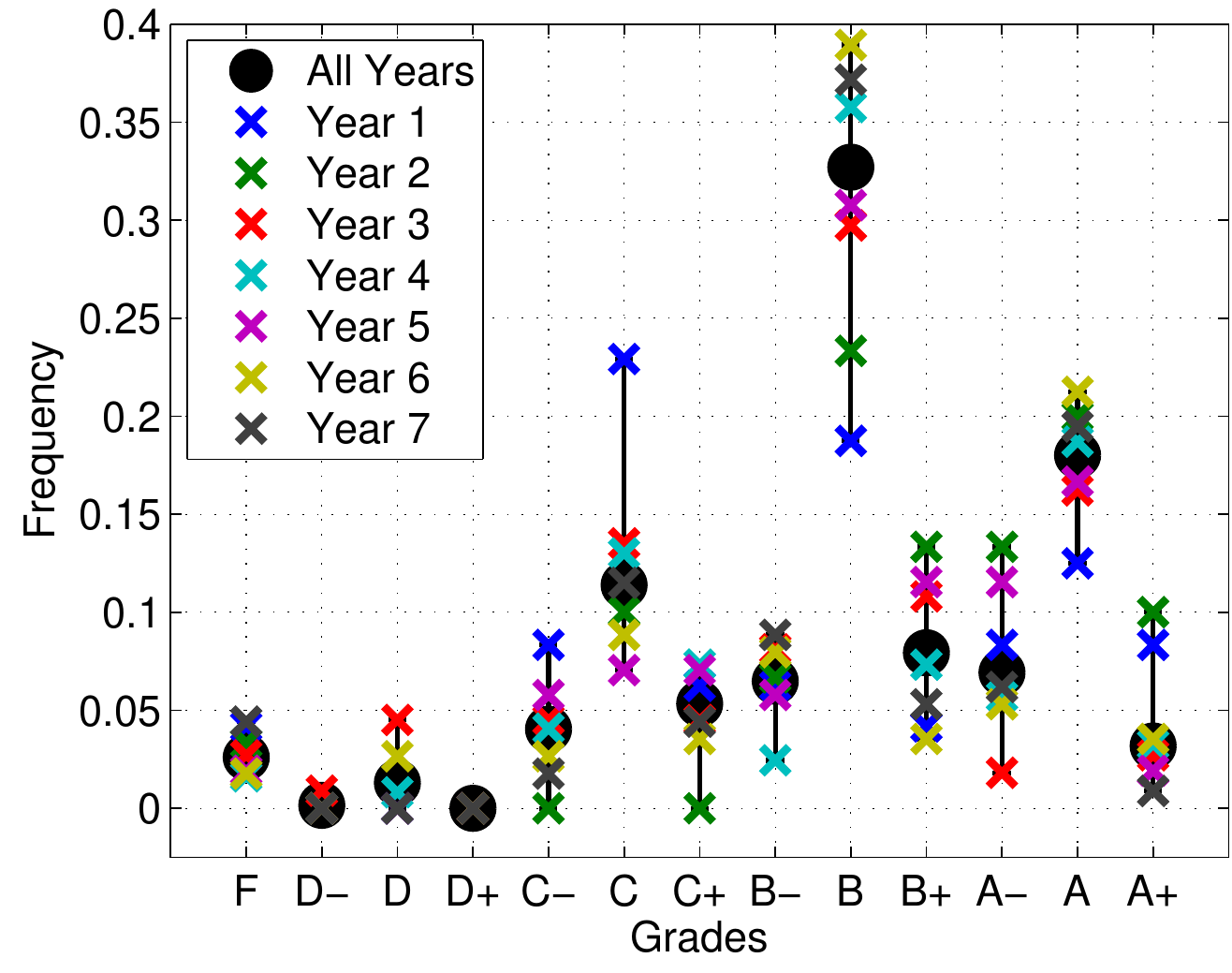}
\label{fig:Grade_Distribution_EE113}}
\hfil
\subfloat[Correlation coefficients overall score]{\includegraphics[width=2.3in]{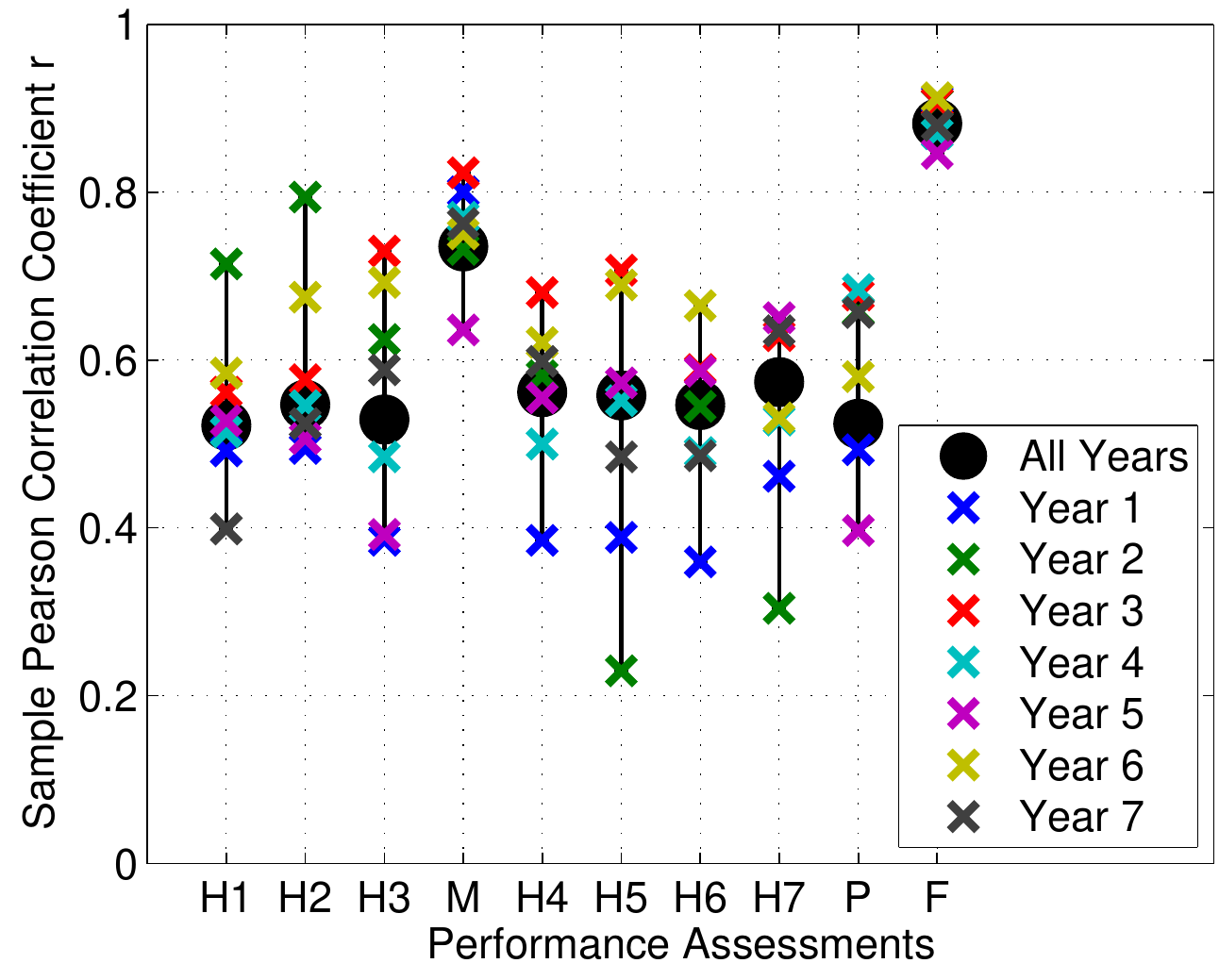}
\label{fig:Correlations_Overall_Score_EE113}}
\hfil
\subfloat[Correlation coefficients final score]{\includegraphics[width=2.3in]{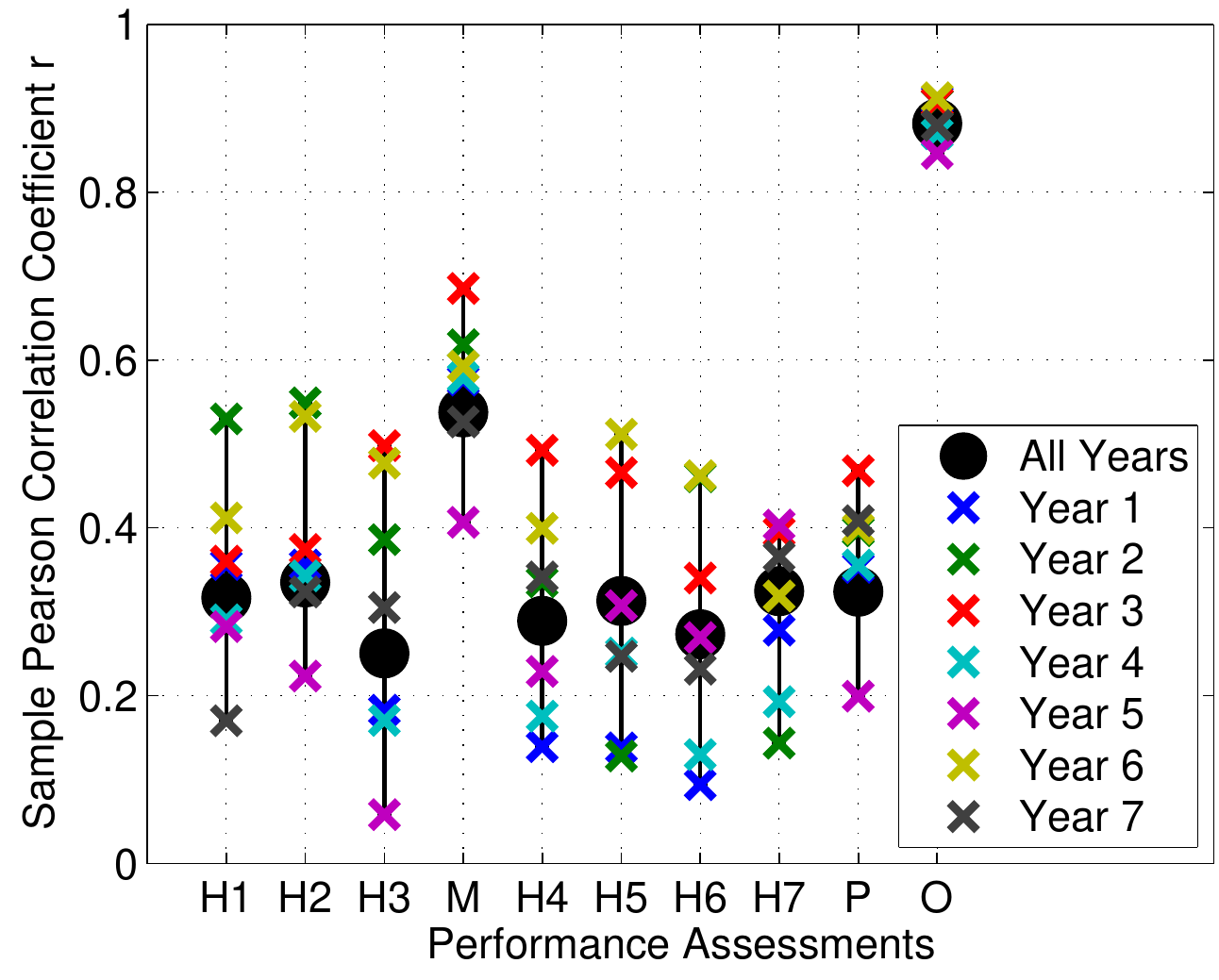}
\label{fig:Correlations_Final_Score_EE113}}
\caption{Data analysis: \ref{fig:Grade_Distribution_EE113} shows the distribution of letter grades for all years. \ref{fig:Correlations_Overall_Score_EE113} and \ref{fig:Correlations_Final_Score_EE113} present the sample Pearson correlation coefficient between individual performance assessments and the overall (\ref{fig:Correlations_Overall_Score_EE113}) or final exam (\ref{fig:Correlations_Final_Score_EE113}) score. Note that we use the abbreviations H$i$ (homework assignment $i$), M (midterm exam), F (final exam) and O (overall score) in the figures.}
\label{fig:Data_Analysis}
\end{figure*}

Our experiments are based on a dataset from an undergraduate digital signal processing course (EE113) taught at UCLA over the past 7 years. The dataset contains the scores from all performance assessments of all students and their final letter grades. The number of students enrolled in the course for a given year varied between $30$ and $156$, in total the dataset contains the scores of approximately $700$ students. Each year the course consists of 7 homework assignments, one in-class midterm exam taking place after the third homework assignment, one course project that has to be handed in after homework 7 and the final exam. The duration of the course is 10 weeks and in each week one performance assessments takes place. The weights of the performance assessments are given by: $20\%$ homework assignments with equal weight on each assignment, $25\%$ midterm exam, $15\%$ course project and $40\%$ final exam.\footnote{As we explain in footnote \ref{ftn:assumption_about_courses} and show in section \ref{subsubsec:learning}, our algorithm can also be applied to settings where the number and weights of performance assessments change over the years.} Fig. \ref{fig:Grade_Distribution_EE113} shows the distribution of the letter grades assigned over the $7$ years. We observe that on average $B$ is the grade the instructor assigned most frequently. $A$ was assigned second most and $C$ third most frequently. Surprisingly, however, the distribution varies drastically over the years; in year $1$ for example only $18.75\%$ received a $B$ while in year 6 the frequency was $38.9\%$.

%\begin{figure}[!t]
	%\centering
		%\includegraphics[width=3in]{Figures/Grade_Distribution_EE113.pdf}
	%\caption{Distribution of letter grades for all years.}
	%\label{fig:Grade_Distribution_EE113}
%\end{figure}

To understand the predictive power of the scores in different performance assessments, Fig. \ref{fig:Correlations_Overall_Score_EE113} shows the sample Pearson correlation coefficient between all performance assessments and the overall score. We make several important observations from this graph. First, on average the final exam has the strongest correlation to the overall score, followed by the midterm exam. This is not surprising, since the final contributes $40\%$ and the midterm contributes $25\%$ to the overall score. Second, the score from the course project on average does not have a higher correlation with the overall score than the homework assignments despite the fact that it accounts for $15\%$ of the overall score. Third, all homework assignments have similar correlation coefficients. Fourth, the correlation between the individual performance assessments and the overall score varies greatly over the years. This indicates that predicting student scores based on training data from past years might be difficult.

%\begin{figure}[!t]
	%\centering
		%\includegraphics[width=3in]{Figures/Correlations_overall_EE113.pdf}
	%\caption{Sample Pearson correlation coefficient between individual performance assessments and the overall score. Note that we use the abbreviations H$i$ (homework assignment $i$), M (midterm exam) and F (final exam) in the figure.}
	%\label{fig:Correlations_Overall_Score_EE113}
%\end{figure}

Since all performance assessments are part of the overall score and, therefore, a high correlation is expected, it is also informative to consider the correlation between the performance assessments and the final exam shown in Fig. \ref{fig:Correlations_Final_Score_EE113}. It is interesting to observe that still the midterm exam shows, besides the overall score, the highest correlation with the final exam. A possible explanation for this is that both the midterm and final are in-class exams while the other performance assessments are take-home.

%\begin{figure}[!t]
	%\centering
		%\includegraphics[width=3in]{Figures/Correlations_final_EE113.pdf}
	%\caption{Sample Person correlation coefficient between individual performance assessments and the final exam. Note that we use the abbreviations H$i$ (homework assignment $i$), M (midterm exam) and O (overall score) in the figure.}
	%\label{fig:Correlations_Final_Score_EE113}
%\end{figure}

\subsection{Our Algorithm}
\label{subsec:implementation_alg}

In this section we discuss four important details of the application of algorithm \ref{alg:cagp_variance} to the dataset from the undergraduate digital signal processing course.

First, the rule we use to normalize all scores $a_{i,y,k}$ in our dataset is given by
\begin{equation}
a_{i,y,k} = \frac{a^*_{i,y,k} - \hat{\mu}_{y,k}}{\hat{\sigma}_{y}},
\label{eq:normalization}
\end{equation}
where $a^*_{i,y,k}$ is the original score of the student, $\hat{\mu}_{y,k}$ is the sample mean of all $y$th year student's original scores in performance assessment $k$ and $\hat{\sigma}_{y}$ is the standard deviation of all $y$th year student's original overall scores. A normalization of the scores is needed for several reasons. First, the instructor-defined maximum score in a particular performance assessment may differ greatly across years and since we use data from past years to predict the performance of students in a given year, we need to make the data across years comparable. Second, also the difficulty of individual performance assessments might be different across years, homework $2$ might for instance be very easy in year $2$ so that almost everyone achieves the maximum score and very difficult in year $3$ so that few achieve half of the maximum score. The normalization according to (\ref{eq:normalization}) eliminates this bias by transforming the absolute scores of a student to scores relative to his/her classmates of the same year. Note that algorithm \ref{alg:cagp_variance} does not require a specific normalization and it does not matter that the normalized scores according to (\ref{eq:normalization}) will not be in the interval $\left[0,1\right]$ as assumed in Section \ref{sec:formalism_algorithm} for simplicity.

Second, we use feature vectors that simply contain the scores of all performance assessments student $i$ has taken up to time $k$ in the order they occurred $\mathbf{x}_{i,y,k}=(a_{i,y,1},\ldots,a_{i,y,k})$. To incorporate the fact that students who have performed similarly in a performance assessment with a lot of weight should be nearer to each other in the feature space than students that have had similar scores in a performance assessment (e.g. homework assignment) with low weight, we use a weighted metric to calculate the distance between two feature vectors. We define the distance of two feature vectors $\mathbf{x}_i,\mathbf{x}_j \in \mathbf{X}^k$ as 
\begin{equation}
\langle\mathbf{x}_i,\mathbf{x}_j\rangle_k = \frac{\sum_{l=1}^k w_l \left| x_{i,l} - x_{j,l} \right|}{\sum_{l=1}^{k} w_l},
\label{eq:distance_metric}
\end{equation}
where $k$ is the length of the feature vectors, $w_l$ is the weight of performance assessment $l$ and $x_{i,l}$ denotes entry $l$ of feature vector $\mathbf{x}_i$.

Third, rather than specifying the radii of the neighborhoods to consider as an input, as suggested in the pseudo-code of algorithm \ref{alg:cagp_variance}, we automatically adapt the radii of the neighborhoods such that they contain a certain number of neighbors. Since the sample variance gets more accurate with an increasing number of samples, we refrain from considering neighborhoods with only $2$ neighbors. Therefore, the smallest radius considered $r_1$ is the minimal radius such that the neighborhood includes $3$ neighbors. For subsequent neighborhoods the minimal radius is chosen such that the neighborhood includes at least one neighbor more than the previous neighborhood. Formally, we define the selection of the radii recursively as
\begin{equation}
\begin{aligned}
r_1 &=\min r,\; s.t. \; |B(\mathbf{x}_{i,y,k},r)| \geq 3 \\
r_{m+1} &= \min r,\; s.t. \; |B(\mathbf{x}_{i,y,k},r)|>|B(\mathbf{x}_{i,y,k},r_m)|.
\end{aligned}
\label{eq:radius_variance}
\end{equation}

Fourth, to be able to apply our algorithm in settings in which structure of the course (e.g. the number, weight and sequence of assessments) changes across years, we need to pre-process the data from past years. In particular, the data from past years is pre-processed so that the number and sequence of in-class and take-home assessments is the same as in the current year. In addition, to identify the most similar students, it is important that the performance assessments that cover the same topic are at the same place of the sequence/feature vector and, therefore, are compared to each other. Consequently, we might have to pre-process the data from past years even if the total number of performance assessments is the same. We use two different types of modifications to pre-process data from past years.

Modification 1 applies to cases where a topic of the course was tested with a larger number of performance assessments in a past year than in a current year. For example, consider a signal processing course which contained two homework assignments on the Fast Fourier Transform in year 1 but the same topic was covered in only one homework assignment in year 2. In this case, the two performance assessments on the same topic from year 1 are combined to a single assessment. If $N$ assessments are combined, the score of the combined assessment $a_{comb}$ is calculated based on the weights $w_1,\ldots,w_N$ of the past assessments with scores $a_1,\ldots,a_N$ according to 
\begin{equation}
a_{comb}=\frac{\sum_{k=1}^N w_k a_k}{ \sum_{k=1}^N w_k}
\end{equation}

Modification 2 applies to the case where a topic of the course was tested with a lower number of performance assessments in a past year than in a current year. In this case the past-year performance assessment on this topic is duplicated. Note that through duplication this performance assessment gets more weight in the process of selecting similar students. This is desired because the instructor probably uses more performance assessments to test a certain topic because he thinks that this topic is very important and hence it will be informative in terms of predicting the grade.

Finally, if necessary the sequence of performance assessments from the past years is reordered to match the sequence of performance assessments from the current year. The reordering has to be done so that the performance assessments on the same topic are at the same position of the sequence/feature vector in both years. Additionally, in-class assessments should only be compared to in-class assessments and take-home assessments should only be compared to take-home assessments. After this pre-processing of past-year data, the standard Algorithm \ref{alg:cagp_variance} can be applied to make the predictions. Note that our algorithm always uses the weights of the current-year course to find students similar to the student for whom it needs to issue a personalized grade prediction and does not consider the weights that were used in past-year courses for the various assessments. The grade predictions for the current-year course are usually made based on data from several past-year courses. The data for each of the past years might have to be pre-processed separately.

\subsection{Benchmarks}
We compare the performance of our algorithm against five different prediction methods.

\begin{itemize}
	\item We use the score $a_{i,y,k}$ student $i$ has achieved in the most recent performance assessment $k$ alone to predict the overall grade.
	
	\item A second simple benchmark makes the prediction based on the scores $a_{i,y,1},\ldots,a_{i,y,k}$ student $i$ has achieved up to performance assessment $k$ taking into account the corresponding weights of the performance assessments.
	
	\item The $k$-Nearest Neighbors algorithm with 7 neighbors. This number provided the best results with training data from the first year.
	
	\item Linear regression using the ordinary least squares (OLS) finds the least squares optimal linear mapping between the scores of first $k$ performance assessments and the overall score.
	
\item In classification settings we use logistic regression instead of linear regression. 

\item Support vector machines (SVMs) are used in the classification setting.

\end{itemize}

The advantage of the method we use in our algorithm over linear and logistic regression is that being a nearest neighbor method, it is able to recognize certain patterns such as trends in the data that are missed in linear/logistic regression where a single parameter per performance assessment has to fit all students. In contrast, our algorithm is able to detect such patterns if there have been students in the past who have shown similar patterns.

Table \ref{tab:CaseStudy} illustrates this through a case study extracted from the UCLA undergraduate digital signal processing course data.
\begin{table}[!t]
	\caption{Case Study: Illustrative Example}
	\centering
		\begin{tabular}{ c | c | c | c | c | c }
		
									& HW 1		& HW 2 		& HW 3		& Midterm 	& Do Well/Poorly	\\
		\hline			
		 Student 1		& 0.53 		&  0.00 	&  -0.37 	&  -1.35 		& Poorly					\\
		\hline					
		 Student 2		& 1.07		& 0.87		&  -0.30 	&  -1.06 		& Poorly					\\
		\hline					
		 Student 3		& -1.39		&-1.54  	& -2.15	 	&   0.50 		& Well      			\\

	\end{tabular}
	\label{tab:CaseStudy}
\end{table}
We present cases from a simulation where we predicted whether students are going to do well (letter grade $\geq B-$) or do poorly (letter grade $\leq C+$) and consider the students for which our algorithm decided to predict after the midterm exam. The table shows 3 students whom logistic regression classified wrongly while our algorithm made the accurate prediction. In columns 2-4 we present the scores the students achieved up to the midterm exam and the last column shows the true classification of the students. These cases are typical examples of settings where our algorithm outperforms logistic regression. Student 1 and 2 both showed a good performance in homework assignment 1. However, in later assignments and especially at the midterm exam their performance successively deteriorated, an indication that the students might do poorly the class if they or the instructor and teaching assistants do not take corrective actions. Our algorithm is likely to have learned such patterns from past data and predicts the students to do poorly. On average, however, their performance in the first four performance assessments is still about average and, therefore, logistic regression predicts that the students will do well. For student 3 the situation is the other way around. 

\subsection{Results}
\label{subsec:results}

In this section we evaluate the performance of our algorithm \ref{alg:cagp_variance} in different settings and compared to benchmarks in both regression and the classification tasks.

As a performance measure in the regression setting, we use the average of the absolute values of the prediction errors $E$. Since we normalized the overall score to have zero mean and a standard deviation of $1$, $E$ directly corresponds to the number of standard deviations the predictions on average are away from the true values. The overall performance measure in classification settings is the accuracy of the classification. Furthermore, we use the quantities precision, recall and false positive/negative rate besides accuracy to measure performance. Please note that positive in our case means that the student does poorly.

\subsubsection{Performance Comparison with Benchmarks in Regression Setting}
\label{subsubsec:results_regression_setting}

\begin{figure}[!t]
	\centering
		\includegraphics[width=3in]{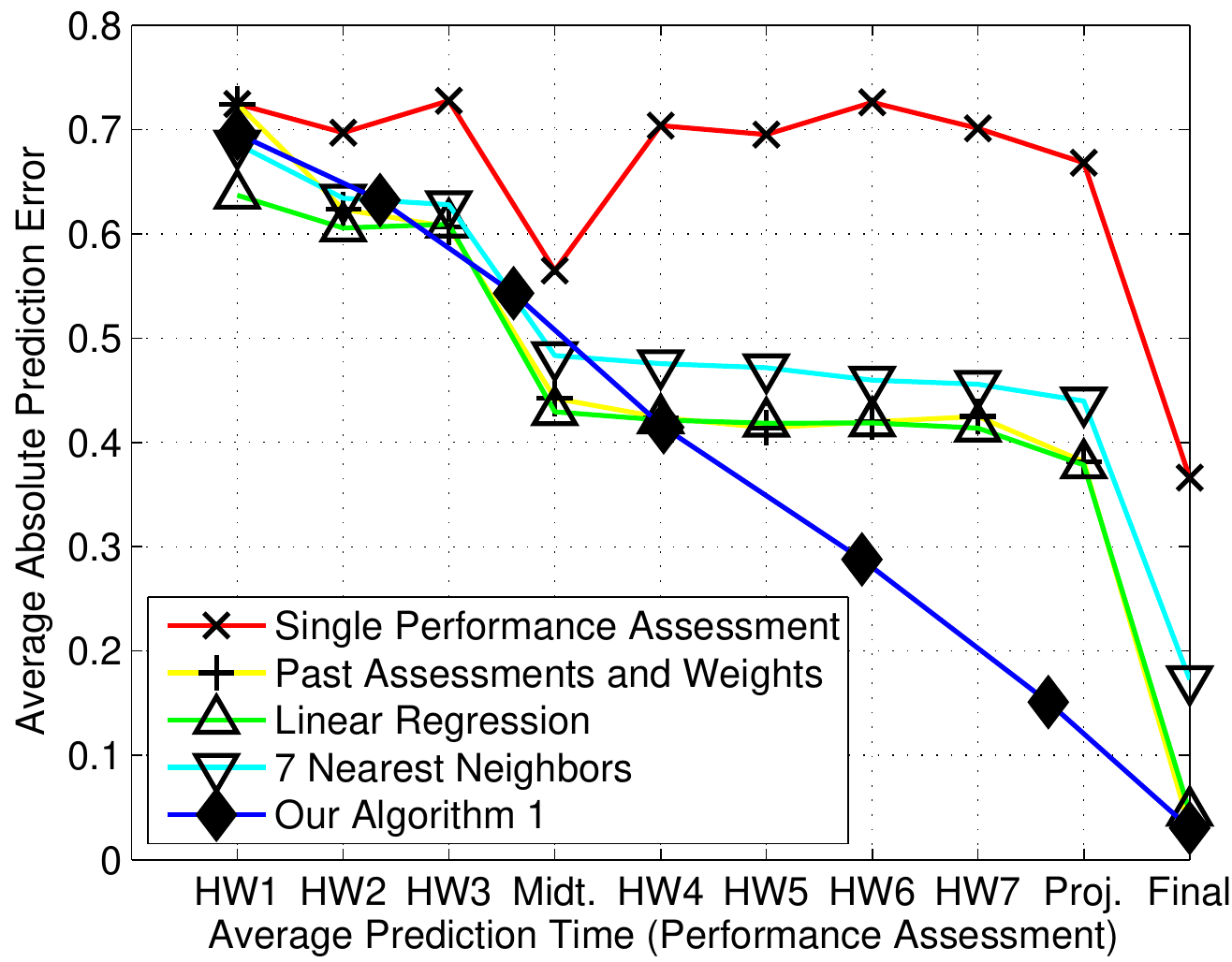}
	\caption{Performance comparison of different prediction methods.}
	\label{fig:Comparison_methods}
\end{figure}

Having discussed the various performance measures, we first address the regression setting. Fig. \ref{fig:Comparison_methods} visualizes the performance of the algorithm we presented in Section \ref{subsec:alg_regression} and of benchmark methods.
%\begin{figure*}[!t]
%\centering
%\subfloat[Performance across different methods]{\includegraphics[width=2.3in]{Figures/Comparison_Methods_Regression.pdf}
%\label{fig:Comparison_methods}}
%\hfil
%\subfloat[Visualization learning]{\includegraphics[width=2.3in]{Figures/Visualization_Learning.pdf}
%\label{fig:Visualization_Learning}}
%\hfil
%\subfloat[Cross instructor predictions]{\includegraphics[width=2.3in]{Figures/Cross_Instructor_Prediction.pdf}
%\label{fig:Cross_Instructor_Prediction}}
%\caption{\ref{fig:Comparison_methods}: Performance comparison of different prediction methods. \ref{fig:Visualization_Learning}: Illustration of learning from past data; error of grade predictions for year 7 depending on training data. \ref{fig:Visualization_Learning}: Illustration of learning across instructors; error of grade predictions for year 7 depending on training data.}
%\end{figure*}
We generated Fig. \ref{fig:Comparison_methods} by predicting the overall scores of all students from years $2-7$. To make the prediction for year $y$, we used the entire data from years $1$ to $y-1$ to learn from. Unlike our algorithms, the benchmark methods do not provide conditions to decide after which performance assessment the decision should be made. Therefore, for benchmark methods we specified the prediction time (performance assessment) $k$ for an entire simulation and repeated the experiment for all $k=1,\ldots,10$; the results are plotted in Fig. \ref{fig:Comparison_methods}. To generate the curve of our algorithm \ref{alg:cagp_variance}, we ran simulations using different confidence thresholds $q_{th}$ and for each threshold we determined $E$ and the performance assessment (time) $\bar{k}$ after which the prediction was made on average.

Irrespective of the prediction method, Fig. \ref{fig:Comparison_methods} shows the trade-off between timeliness and accuracy; the later we predict the more accurate our prediction gets. From the curve for the prediction using a single performance assessment we infer that there is a low correlation between homework assignments/course project and the overall score and a high correlation between the in-class assessments (midterm and final exam) and the overall score. This observation is congruent with the correlation analysis from Section \ref{subsec:dataAnalysis}. If the prediction is made early, before the midterm, all methods (except the prediction using a single performance assessment) lead to similar prediction errors. We observe that while the error decreases approximately linearly for our algorithm, the performance of benchmark methods steeply increases after the midterm and the final but stays approximately constant during the rest of the time. The reason for this is that we obtained the points of the curve for our algorithm by averaging the prediction time of all students. Therefore, the point of the curve above the midterm was not generated by predicting after the midterm for all students; some predictions were made earlier, some later. If on average the prediction is made after homework $4$, our algorithm shows a significantly smaller error $E$ than benchmark methods outperforming linear regression by up to $65\%$.

\subsubsection{Learning across Years and Instructors in Regression Setting}
\label{subsubsec:learning}

\begin{figure}[!t]
	\centering
		\includegraphics[width=3in]{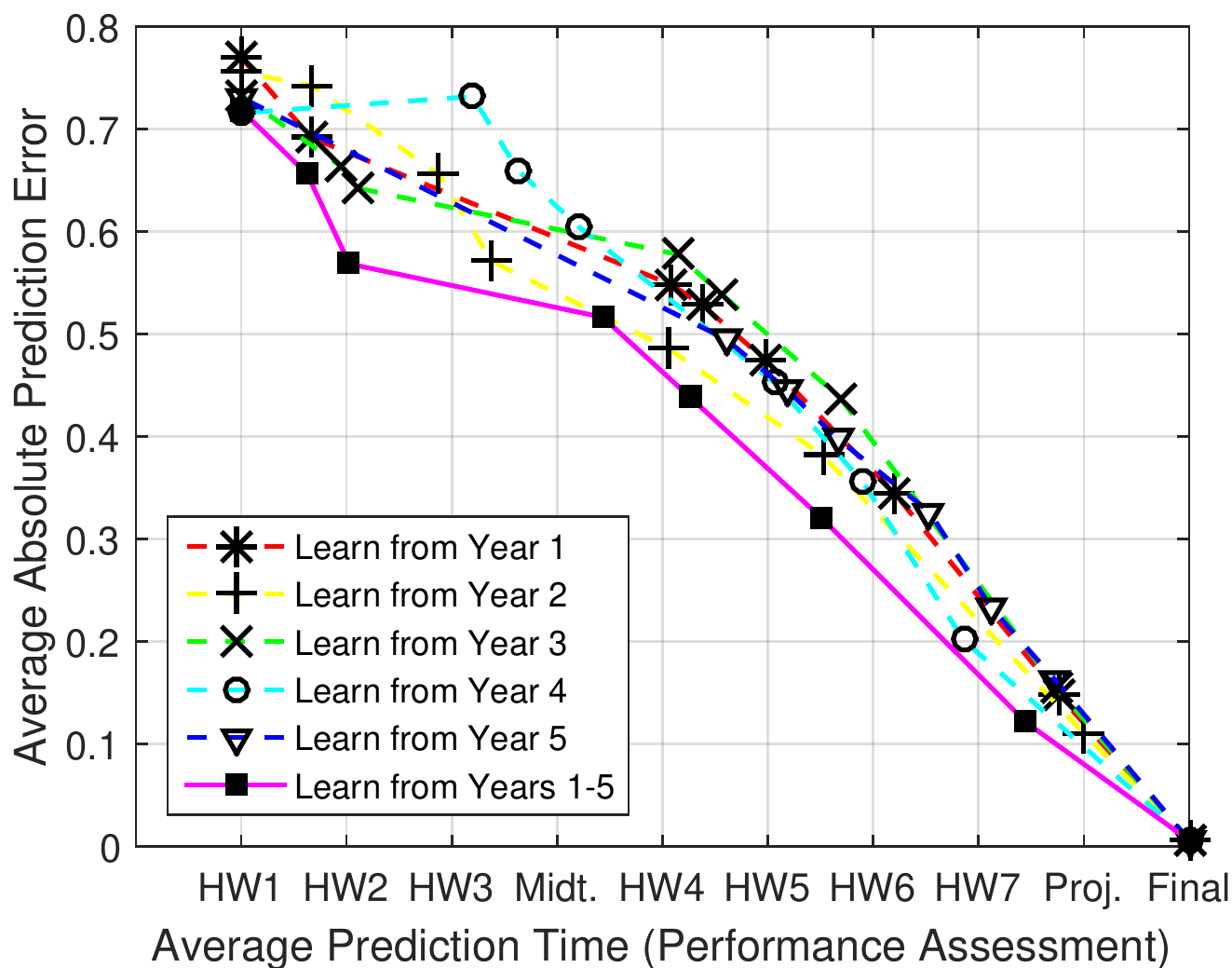}
	\caption{Illustration of learning from past data: Error of grade predictions for year 7 depending on training data.}
	\label{fig:Visualization_Learning}
\end{figure}

Consider Fig. \ref{fig:Visualization_Learning} demonstrating the performance increase of our algorithm when more data to learn from become available.
To generate the figure, we used our algorithm to predict the overall scores of all 7th year students for different confidence thresholds. The curves in dashed lines stem from simulations using only one of the years 1-5 as training data and the solid magenta curve uses all years 1-5 to learn from. We observe that the prediction performance strongly depends on the training data and differs if different years are used. Most importantly, the performance is highest irrespective of the average prediction time if the combination of the data from all 5 years is used. This shows that our algorithm is able to learn and improves its predictions over time.

\begin{figure}[!t]
	\centering
		\includegraphics[width=3in]{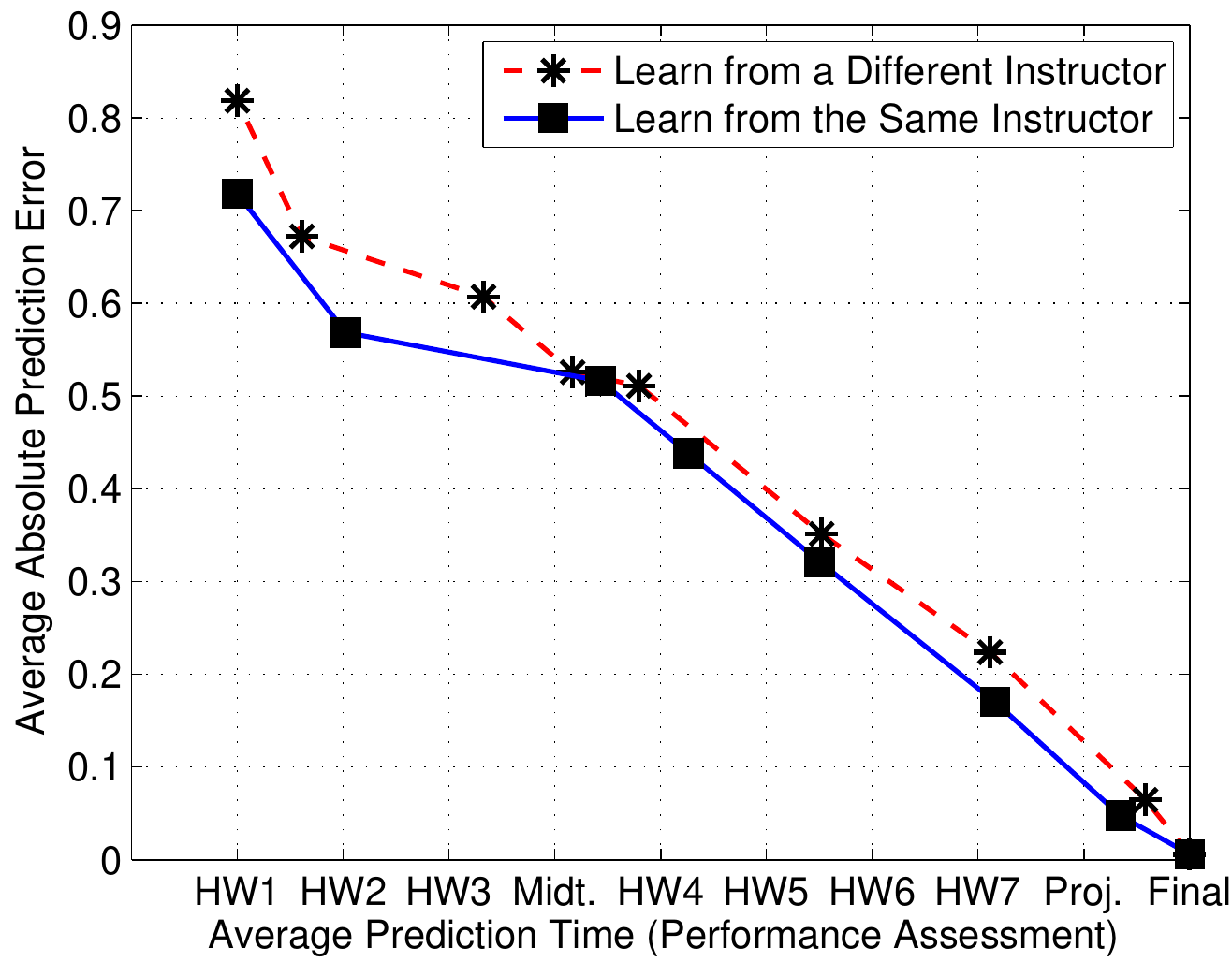}
	\caption{Illustration of learning across instructors: Error of grade predictions for year 7 depending on training data.}
	\label{fig:Cross_Instructor_Prediction}
\end{figure}

The undergraduate digital signal processing course is taught twice a year by three different instructors at UCLA. While we used only the data from one instructor in the previous plots, Fig. \ref{fig:Cross_Instructor_Prediction} investigates the situation when we predict the grades for a class of instructor 1 based exclusively on past data from a different instructor 2.
In practice this happens when a new instructor takes over a course previously taught by someone else. It is interesting to see whether our grade prediction still works well in this setting. A good performance is not self-evident for several reasons. Different instructors might set a different focus concerning the knowledge imparted, they might use a different textbook and they might prefer different styles of homework assignments and in-class exams. Furthermore, the structure of the course, e.g. the number and sequence of homework assignments, the time when the midterm exam takes place, the weights of performance assessments and whether a course project and quizzes exist, might change drastically. To generate Fig. \ref{fig:Cross_Instructor_Prediction}, we predicted the overall score for the year 7 class of instructor 1 based on two different sets of previous data. The solid blue curve was generated by using the data from the classes in years 1-5 from the same instructor 1 as training data. To obtain the dashed red curve, we used the data from classes in years 1-5 from instructor 2 to learn from. While the predictions using training data from the same instructor are slightly more accurate, the performance with training data from a different instructor is still very satisfying, showing a good robustness of our algorithm with respect to different instructors. For the subsequent results we again exclusively use data from one instructor.

\subsubsection{Performance Comparison with Course Containing Early Quizzes in Regression Setting}
\begin{figure}[!t]
	\centering
		\includegraphics[width=3in]{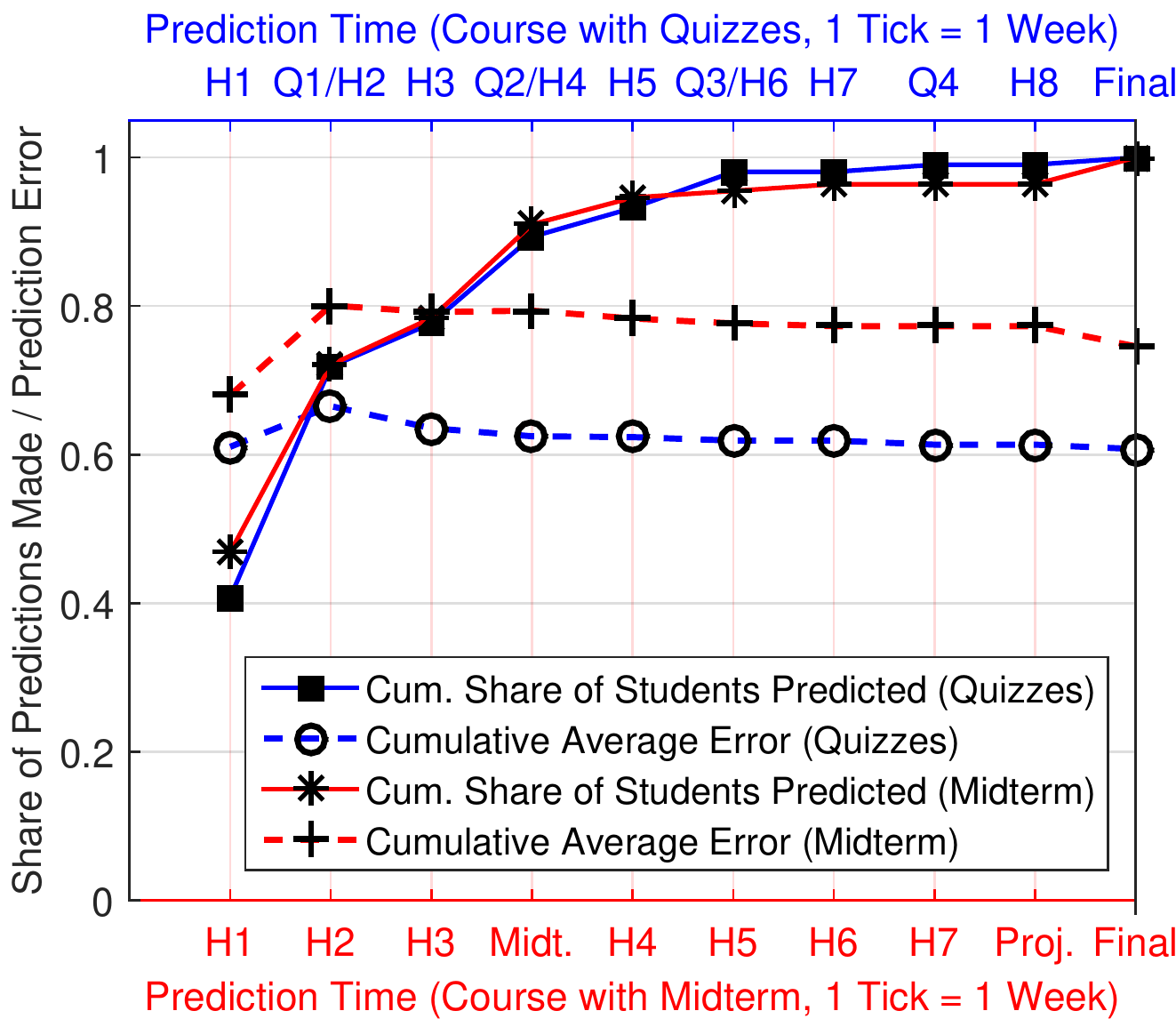}
	\caption{Comparison of prediction time and accuracy between the UCLA course EE113, which contains a midterm exam, and the UCLA course EE103, which contains four in-class quizzes instead of a midterm exam. Note that the tick labels Q$i$/H$i$ above the plot stand for quiz/homework $i$  and that for EE103 there are weeks in which both a homework and a quiz take place.}
	\label{fig:Cumulative_Comparison_Quiz}
\end{figure}

The results in both the data analysis section (Fig. \ref{fig:Correlations_Overall_Score_EE113}) and Section \ref{subsubsec:results_regression_setting} (Fig. \ref{fig:Comparison_methods}) indicate that scores in in-class exams are much better predictors of the overall score than homework assignments. To verify this, we consider two consecutive years of the UCLA course EE103, which contains four in-class quizzes in course weeks 2, 4, 6 and 8 instead of a midterm. Fig. \ref{fig:Cumulative_Comparison_Quiz} visualizes that, starting from the first quiz in week 2, indeed our algorithm is able to predict the same percentage of the students with an up to $22\%$ smaller cumulative average prediction error by a certain week. 
%\begin{figure*}[!t]
%\centering
%\subfloat[Performance across different methods]{\includegraphics[width=2.3in]{Figures/Cumulative_Comparison_Quiz.pdf}
%\label{fig:Cumulative_Comparison_Quiz}}
%\hfil
%\subfloat[Visualization learning]{\includegraphics[width=2.3in]{Figures/Binary_Comparison_Acc_Prec_Rec.pdf}
%\label{fig:Binary_comparison_acc_prec_rec}}
%\hfil
%\subfloat[Cross instructor predictions]{\includegraphics[width=2.3in]{Figures/Time_vs_Cumulative_Accuracy.pdf}
%\label{fig:Time_vs_Cumulative_Accuracy}}
%\caption{\ref{fig:Cumulative_Comparison_Quiz}: Comparison of prediction time and accuracy between the UCLA course EE113, which contains a midterm exam, and the UCLA course EE103, which contains four in-class quizzes instead of a midterm exam. Note that the tick labels Q$i$/H$i$ above the plot stand for quiz/homework $i$  and that for EE103 there are weeks in which both a homework and a quiz take place. \ref{fig:Binary_comparison_acc_prec_rec}: Performance comparison between our algorithm and logistic regression using accuracy, precision and recall for binary do well/poorly classification. \ref{fig:Time_vs_Cumulative_Accuracy}: Cumulative prediction time, accuracy, false positive and false negative error rates for a binary do well/poorly classification with fixed $q_{th}$.}
%\end{figure*}
We generated Fig. \ref{fig:Cumulative_Comparison_Quiz} by using algorithm \ref{alg:cagp_variance} to predict for both courses the overall scores of the students in a particular year based on data from the previous year. Note that for the course with quizzes, the increase in the share of students predicted is larger in weeks that contain quizzes than in weeks without quizzes. This supports the thesis that quizzes are good predictors as well.

According to this result, it is desirable to design courses with early in-class exams. This enables a timely and accurate grade prediction based on which the instructor can intervene if necessary.

\subsubsection{Performance Comparison with Benchmarks in Classification Setting}
\begin{figure}[!t]
	\centering
		\includegraphics[width=3in]{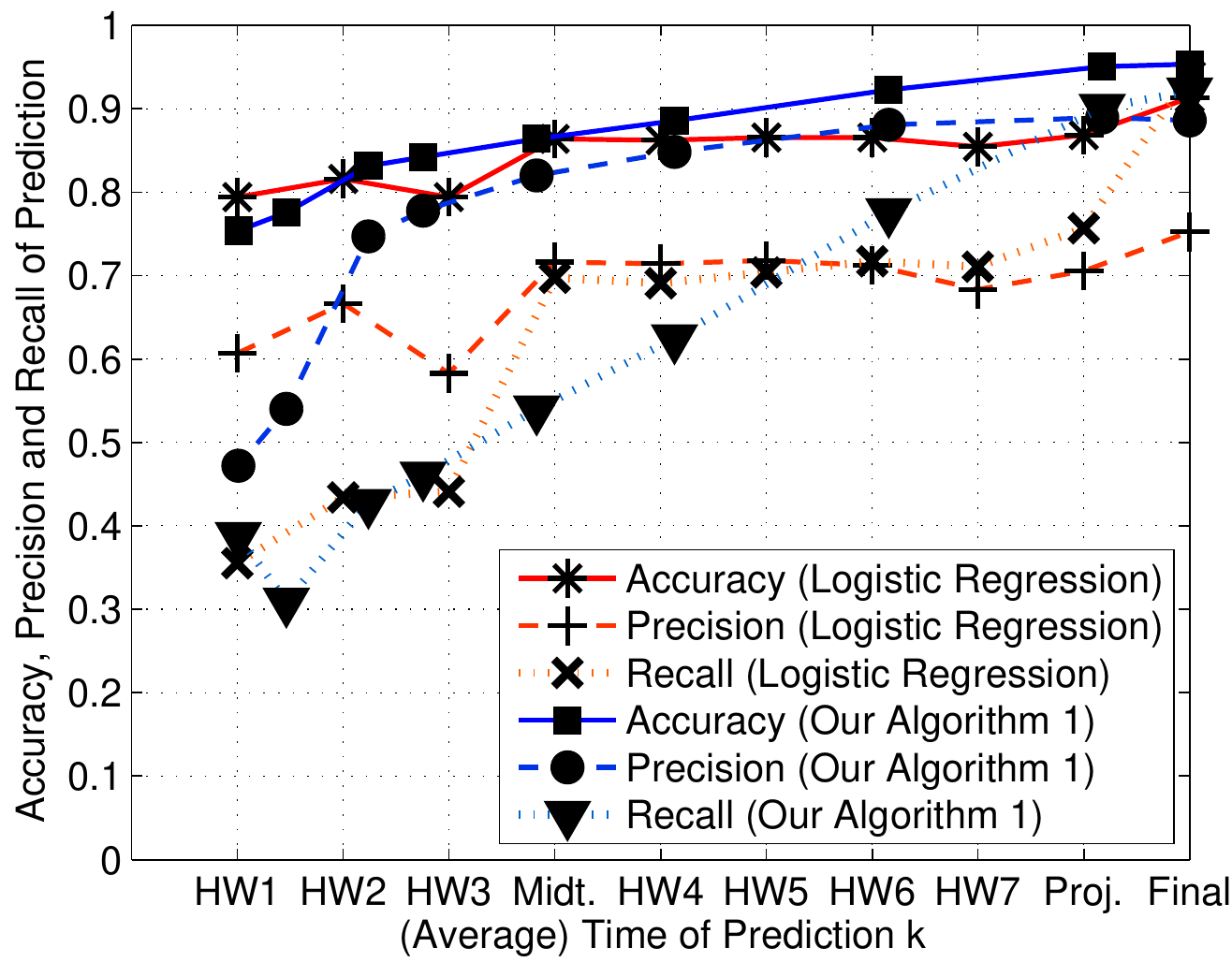}
	\caption{Performance comparison between our algorithm and logistic regression using accuracy, precision and recall for binary do well/poorly classification.}
	\label{fig:Binary_comparison_acc_prec_rec}
\end{figure}

The performances in the binary classification settings are summarized in Fig. \ref{fig:Binary_comparison_acc_prec_rec}.
Since logistic regression turns out to be the most challenging benchmark in terms of accuracy in the classification setting, we do not show the performance of the other benchmark algorithms for the sake of clarity. The goal was to predict whether a student is going to do well, still defined as letter grades equal to or above $B-$, or do poorly, defined as letter grades equal to or below $C+$. Again, to generate the curves for the benchmark method, logistic regression, we specified manually when to predict. For our algorithm we again averaged the prediction times of an entire simulation and varied $q_{th}$ to obtain different points of the curve. Up to homework 4, the performance of the two algorithms is very similar, both showing a high prediction accuracy even with few performance assessments. Starting from homework 4, our algorithm performs significantly better, with an especially drastic improvement of recall. It is interesting to see that even with full information, the algorithms do not achieve a $100\%$ prediction accuracy. The reason for this is that the instructor did not use a strict mapping between overall score and letter grade and the range of overall scores that lead to a particular letter grade changed slightly over the years.

\subsubsection{Decision Time and Accuracy in Classification Setting}
\begin{figure}[!t]
	\centering
		\includegraphics[width=3in]{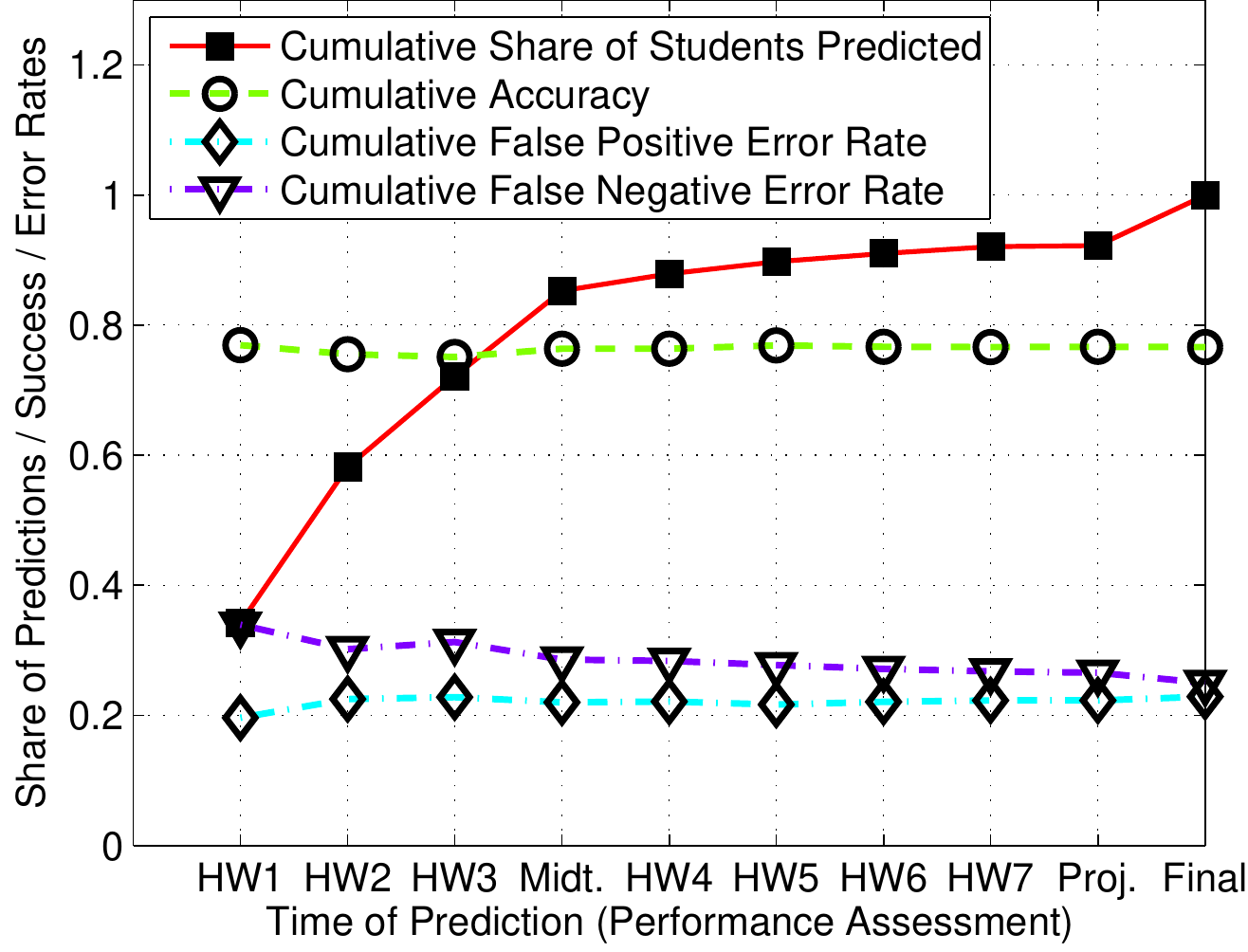}
	\caption{Cumulative prediction time, accuracy, false positive and false negative error rates for a binary do well/poorly classification with fixed $q_{th}$.}
	\label{fig:Time_vs_Cumulative_Accuracy}
\end{figure}

To better understand when our algorithm makes decisions and with what accuracy, consider Fig. \ref{fig:Time_vs_Cumulative_Accuracy}.
We again investigate binary do well/poorly classifications as discussed above. The red curve shows (square markers) for what share of the total number of students the algorithm makes the prediction by a specific point in time. The remaining curves show different measures of cumulative performance. We can for example see that by the midterm exam we classify $85\%$ of the students with an accuracy of $76\%$. These timely predictions are desirable since the earlier the prediction is made the more time an instructor has to take corrective action. The cumulative accuracy stays almost constant around $80\%$ irrespective of the prediction time. We believe that the reason for this is that thanks to the confidence threshold, the easy decisions are made early and harder decisions are made later. Consequently, the expected accuracy of all predictions remains more or less constant irrespective of the prediction time.

\section{Conclusion}
\label{sec:conclusion}

In this paper we develop an algorithm that allows for a timely and personalized prediction of the final grades of students exclusively based on their scores in early performance assessment such as homework assignments, quizzes or midterm exams. Using data from an undergraduate digital signal processing course taught at UCLA, we show that the algorithm is able to learn from past data, that it outperforms benchmark algorithms with regard to accuracy and timeliness both in classification and regression settings and that the predictions are robust even when the course is taught by different instructors.

We show that in-class exams are better predictors of the overall performance of a student than homework assignments. Hence, designing courses to have early in-class evaluations enables timely identification of students who, with a high probability, would do poorly without intervention and enables remedial actions to be adopted at an early stage.

Our algorithm can easily be generalized to include context data from students such as their prior GPA or demographic data. If applied exclusively to MOOCs, the in-course data used for the predictions could be extended for example by the responses of students to multiple-choice questions, their forum activity, the course material they studied or the time they spent studying online. Another direction of future work is to apply our algorithm in practice and investigate to what extent the performance of students can be improved by a timely intervention based on the grade predictions. In this context, our algorithm could be extended to make multiple predictions for each student to monitor the trend in the predicted grade after an intervention.

One example for an intervention would be that the instructor provides additional study material to students with a low predicted grade. Alternatively, teaching assistants could spend additional time with these selected students to go through important topics again. In a MOOC setting, the intervention could take place in a fully automated way, for example by presenting the students additional study material in a personalized way using techniques discussed in \cite{tekinetutor}. To make students aware of their performance, they could be asked to predict their own overall grade and as a comparison the instructor could disclose the prediction of our algorithm to the students.

% if have a single appendix:
%\appendix[Proof of the Zonklar Equations]
% or
\appendix  % for no appendix heading

In this Appendix, we proof the theorem from section \ref{subsec:alg_regression}. Before we start with the proof, we discuss some preliminary results.

\begin{fact}
\label{fact:hoeffding_bound}
(Chernoff-Hoeffding Bound) Let $X_1,X_2,\ldots,X_n$ be independent and bounded random variables with range $[0,1]$ and expected value $\mu$. Let $\hat{\mu}_n = (X_1+\ldots+X_n)/n$ denote the sample mean of the random variables. Then, for all $\epsilon > 0$
\begin{equation*}
P\left[ \left| \hat{\mu}_n - \mu \right| \geq \epsilon \right] \leq 2 \exp\left[-2 n \epsilon^2\right].
\end{equation*}
\end{fact}
\begin{IEEEproof}
A proof of Fact \ref{fact:hoeffding_bound} can be found in Hoeffding's paper \cite{hoeffding1963probability}.
\end{IEEEproof}

\begin{fact}
\label{fact:bernstein_bound}
(Empirical Bernstein Bound) Let $n\geq2$ and $X_1,X_2,\ldots,X_n$ be independent and bounded random random variables with range $[0,1]$ and variance $Var$. $\hat{\mu}_n$ denotes the $n$-sample mean $\hat{\mu}_n= \frac{1}{n} \sum_{i=1}^n{X_i}$ and $\widehat{Var}_n$ denotes the $n$-sample variance $\widehat{Var}_n = \frac{1}{n-1} \sum_{i=1}^N \left(x_i - \hat{\mu}_n\right)^2$. Then, the following inequality bounds the probability that the error of the sample standard deviation, which is the square root of the sample variance, is larger than a given value
\begin{equation*}
P\left[\left| \sqrt{\widehat{Var}_n} - \sqrt{Var} \right| \geq \epsilon  \right] \leq 2 \exp\left[ -\frac{n-1}{2} \epsilon^2 \right]
\end{equation*}
can be derived.
\end{fact}
\begin{IEEEproof}
See \cite{maurer2009empirical} for a proof of Fact \ref{fact:bernstein_bound}.
\end{IEEEproof}

\begin{lemma}
\label{lemma:wrong_ball}
Let $\hat{m}_k(\mathbf{x})$ denote the index of the neighborhood selected by our algorithm for the student with feature vector $\mathbf{x}$ at time $k$ and $m_k^*(\mathbf{x})$ is given by (\ref{eq:def_hood_smallest_var}). $M$ denotes the total number of neighborhoods our algorithm considers and $\Delta_k(\mathbf{x})$ is given by (\ref{eq:Delta_definition}). We can bound the probability that our algorithm chooses the wrong neighborhood by
\begin{equation}
P\left[ \hat{m}_k(\mathbf{x}) \neq m_k^*(\mathbf{x}) \right] \leq 2 e^{- \Delta_k(\mathbf{x})^2 \underset{1 \leq m \leq M}{\min} \frac{|B(\mathbf{x},r_m)|-1}{8}}.
\end{equation}
\end{lemma}
\begin{IEEEproof}
Consider:
\begin{align*}
& P\left[ \hat{m}_k(\mathbf{x})  \neq m_k^*(\mathbf{x}) \right]\\
& = P\left[ \underset{1 \leq m \leq M}{\argmin} \  \widehat{Var}^k(\mathbf{x},r_m) \neq m_k^*(\mathbf{x}) \right] \\
& = P\left[ \underset{1 \leq m \leq M}{\argmin} \  \sqrt{\widehat{Var}^k(\mathbf{x},r_m)} \neq m_k^*(\mathbf{x}) \right].
\end{align*}
If the estimation error of the standard deviation is smaller than $\Delta_k(\mathbf{x})/2$ for all neighborhoods
\begin{equation*}
\left| \sqrt{\widehat{Var}^k(\mathbf{x},r_m)} - \sqrt{Var^k(\mathbf{x},r_m)} \right| \leq \frac{\Delta_k(\mathbf{x})}{2},
\end{equation*}
our algorithm chooses the optimal neighborhood $m_k^*(\mathbf{x})$. Therefore, we get
\begin{align*}
& P\left[ \underset{1 \leq m \leq M}{\argmin} \  \sqrt{\widehat{Var}^k(\mathbf{x},r_m)} \neq m_k^*(\mathbf{x}) \right] \\
& \begin{aligned} \leq P\left[ \underset{1 \leq m \leq M}{\bigcup} \right. & \left\{ \left| \sqrt{\widehat{Var}^k(\mathbf{x},r_m)}  \right. \right.  \\
& \left. \left. \left. - \sqrt{Var^k(\mathbf{x},r_m)} \right|   \geq \frac{\Delta_k(\mathbf{x})}{2}  \right\} \right] \end{aligned}\\
& \overset{(a)}{\leq} \sum_{m=1}^{M} P\left[ \left| \sqrt{\widehat{Var}^k(\mathbf{x},r_m)} - \sqrt{Var^k(\mathbf{x},r_m)} \right| \geq \frac{\Delta_k(\mathbf{x})}{2}    \right] \\
& \overset{(b)}{\leq} \sum_{m=1}^{M} 2 \exp\left[- \Delta_k(\mathbf{x})^2 \frac{|B(\mathbf{x},r_m)|-1}{8}\right] \\
& \leq 2 M \exp\left[- \Delta_k(\mathbf{x})^2 \underset{1 \leq m \leq M}{\min} \frac{|B(\mathbf{x},r_m)|-1}{8}\right]
\end{align*}
where ($a$) is the union bound and ($b$) follows from Fact \ref{fact:bernstein_bound}.
\end{IEEEproof}

\begin{IEEEproof}[Proof of Theorem]
Note that
\begin{align*}
& \left| z_{i,y} - \hat{z}_{i,y,k} \right| \\
& \overset{(a)}{=} \left| c_{i,y,k} + \sum_{l=1}^k {w_l a_{i,y,l}} - \left( \hat{c}_{i,y,k} + \sum_{l=1}^k {w_l a_{i,y,l}} \right) \right| \\
& = \left| c_{i,y,k} - \hat{c}_{i,y,k} \right|
\end{align*}
where ($a$) follows from equations (\ref{eq:overall_residual}) and (\ref{eq:est_overall_score}).

There are three sources of error in the prediction of an overall score of algorithm \ref{alg:cagp_variance}:
\begin{enumerate}
	\item The wrong neighborhood size may be selected due to inaccurate approximations of the true residual score variances of the neighborhoods through the sample variance.
	\item If the optimal neighborhood is selected, the sample mean of the residual scores in the neighborhood may not be a good approximation of their true mean.
	\item Even if the optimal neighborhood is selected and the sample mean equals the true mean, the residual score of the considered student may be different from the mean of the residual score distribution.
\end{enumerate}
In the following we separate these three error sources and derive a bound for each one.

We have
\begin{align*}
 P & \left[ \left| z_{i,y} - \hat{z}_{i,y,k} \right|  \geq \epsilon \right] =  P \left[ \left| c_{i,y,k} - \hat{c}_{i,y,k} \right|  \geq \epsilon \right] \\
\overset{(b)}{=}  & P \left[ \left| c_{i,y,k} - \hat{\mu}^k\left(\mathbf{x},r_{\hat{m}_k(\mathbf{x})}\right) \right|  \geq \epsilon \right]  \\
\overset{(c)}{=} & P\left[ \left| c_{i,y,k} - \hat{\mu}^k\left(\mathbf{x},r_{\hat{m}_k(\mathbf{x})}\right) \right|  \geq \epsilon, \hat{m}_k(\mathbf{x}) = m^*_k(\mathbf{x}) \right] \\
		& + P\left[ \left| c_{i,y,k} - \hat{\mu}^k\left(\mathbf{x},r_{\hat{m}_k(\mathbf{x})}\right) \right|  \geq \epsilon, \hat{m}_k(\mathbf{x}) \neq m^*_k(\mathbf{x}) \right] \\
\overset{(d)}{\leq} & P\left[ \left| c_{i,y,k} - \hat{\mu}^k\left(\mathbf{x},r_{m^*_k(\mathbf{x})}\right) \right|  \geq \epsilon, \hat{m}_k(\mathbf{x}) = m^*_k(\mathbf{x}) \right] \\
		& + P\left[ \hat{m}_k(\mathbf{x}) \neq m^*_k(\mathbf{x}) \right] \\
\overset{(e)}{\leq} & P\left[ \left| c_{i,y,k} - \hat{\mu}^k\left(\mathbf{x},r_{m^*_k(\mathbf{x})}\right) \right|  \geq \epsilon \right] + P\left[ \hat{m}_k(\mathbf{x}) \neq m^*_k(\mathbf{x}) \right]
\end{align*}
where (b) follows from (\ref{eq:notation_best_residual}), (c) is the law of total probability and (d) and (e) both follow from the fact that $P[A,B]\leq P[A]$. Lemma \ref{lemma:wrong_ball} provides a bound for the second term. Therefore, we focus on the first term
\begin{align*}
 & P \left[ \left| c_{i,y,k} - \hat{\mu}^k\left(\mathbf{x},r_{m^*_k(\mathbf{x})}\right) \right|  \geq \epsilon \right] \\
 &  \begin{aligned} = P\left[ \left| c_{i,y,k} - \mu^k\left(\mathbf{x},r_{m_k^*(\mathbf{x})}\right) \right. \right. & \left. \left. + \mu^k\left(\mathbf{x},r_{m_k^*(\mathbf{x})}\right) \right. \right. \\
		& \left. \left. - \hat{\mu}^k\left(\mathbf{x},r_{m^*_k(\mathbf{x})}\right) \right|  \geq \epsilon \right] \end{aligned} \\
& \overset{(f)}{\leq}  P\left[ \left| c_{i,y,k} - \mu^k\left(\mathbf{x},r_{m_k^*(\mathbf{x})}\right) \right| \geq \frac{\epsilon}{2} \right] \\
& \quad  + P\left[ \left| \mu^k\left(\mathbf{x},r_{m_k^*(\mathbf{x})}\right) - \hat{\mu}^k\left(\mathbf{x},r_{m^*_k(\mathbf{x})}\right) \right| \geq \frac{\epsilon}{2} \right] \\
& \overset{(g)}{\leq} \frac{4 Var^k\left(\mathbf{x},r_{m_k^*(\mathbf{x})}\right)}{\epsilon^2} + 2 \exp\left[-\frac{\epsilon^2}{2} \left|B\left(\mathbf{x},r_{m_k^*(\mathbf{x})}\right)\right|\right] \\
& \leq \frac{4 Var^k\left(\mathbf{x},r_{m_k^*(\mathbf{x})}\right)}{\epsilon^2} + 2 \exp\left[-\epsilon^2 \underset{1 \leq m \leq M}{\min} \frac{\left|B\left(\mathbf{x},r_{m}\right)\right|}{2}\right]
\end{align*}
where ($f$) follows from the triangle inequality, the fact that $P[X+Y\geq x_0+y_0] \leq P\left[\left\{X\geq x_0\right\} \cup \left\{Y\geq y_0\right\}\right]$ and the union bound. The bound for the first term in step ($g$) follows from Chebyshev's inequality and the bound for the second term follows from the Chernoff-Hoeffding Bound from Fact \ref{fact:hoeffding_bound}.

Including the second term again and using Lemma \ref{lemma:wrong_ball} we get
\begin{align*}
& P \left[ \left| z_{i,y} - \hat{z}_{i,y,k} \right|  \geq \epsilon \right] =  P \left[ \left| c_{i,y,k} - \hat{c}_{i,y,k} \right|  \geq \epsilon \right] \\
& \leq P\left[ \left| c_{i,y,k} - \hat{\mu}^k\left(\mathbf{x},r_{m^*_k(\mathbf{x})}\right) \right|  \geq \epsilon \right] + P\left[ \hat{m}_k(\mathbf{x}) \neq m^*_k(\mathbf{x}) \right] \\
& \leq \frac{4 Var^k\left(\mathbf{x},r_{m_k^*(\mathbf{x})}\right)}{\epsilon^2} + 2 \exp\left[-\epsilon^2 \underset{1 \leq m \leq M}{\min} \frac{\left|B\left(\mathbf{x},r_{m}\right)\right|}{2}\right] \\
& \quad + 2 M \exp\left[- \Delta_k(\mathbf{x})^2 \underset{1 \leq m \leq M}{\min} \frac{|B(\mathbf{x},r_m)|-1}{8}\right],
\end{align*}
which concludes the proof.
\end{IEEEproof}

% do not use \section anymore after \appendix, only \section*
% is possibly needed

% use appendices with more than one appendix
% then use \section to start each appendix
% you must declare a \section before using any
% \subsection or using \label (\appendices by itself
% starts a section numbered zero.)
%

%
%\appendices
%\section{Proof of the First Zonklar Equation}
%Appendix one text goes here.

% you can choose not to have a title for an appendix
% if you want by leaving the argument blank
%\section{}
%Appendix two text goes here.

% use section* for acknowledgment
%\section*{Acknowledgment}

% Can use something like this to put references on a page
% by themselves when using endfloat and the captionsoff option.
\ifCLASSOPTIONcaptionsoff
  \newpage
\fi

% trigger a \newpage just before the given reference
% number - used to balance the columns on the last page
% adjust value as needed - may need to be readjusted if
% the document is modified later
%\IEEEtriggeratref{8}
% The "triggered" command can be changed if desired:
%\IEEEtriggercmd{\enlargethispage{-5in}}

% references section

% can use a bibliography generated by BibTeX as a .bbl file
% BibTeX documentation can be easily obtained at:
% http://www.ctan.org/tex-archive/biblio/bibtex/contrib/doc/
% The IEEEtran BibTeX style support page is at:
% http://www.michaelshell.org/tex/ieeetran/bibtex/
\bibliographystyle{IEEEtran}
% argument is your BibTeX string definitions and bibliography database(s)
\bibliography{IEEEabrv,bibliography}

% biography section
% 
% If you have an EPS/PDF photo (graphicx package needed) extra braces are
% needed around the contents of the optional argument to biography to prevent
% the LaTeX parser from getting confused when it sees the complicated
% \includegraphics command within an optional argument. (You could create
% your own custom macro containing the \includegraphics command to make things
% simpler here.)
\begin{IEEEbiography}[{\includegraphics[width=1in,height=1.25in,clip,keepaspectratio]{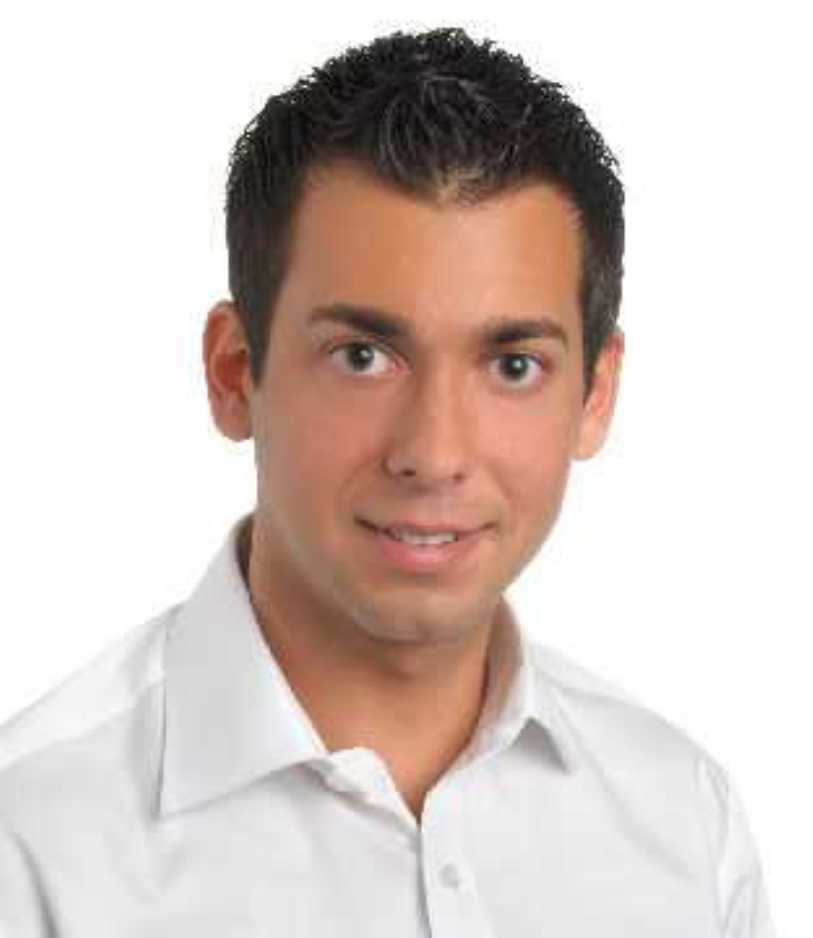}}]{Yannick Meier}
received the B.Sc. degree in information technology and electrical engineering from ETH Zurich, Zurich, Switzerland (Swiss Federal Institute of Technology Zurich) in 2014.

In 2014 and 2015 he conducted research visits at University of Pennsylvania, Philadelpha, PA, USA and at University of California, Los Angeles, CA, USA. He is currently pursuing the M.Sc. degree in information technology and electrical engineering at ETH Zurich.
\end{IEEEbiography}

\begin{IEEEbiography}[{\includegraphics[width=1in,height=1.25in,clip,keepaspectratio]{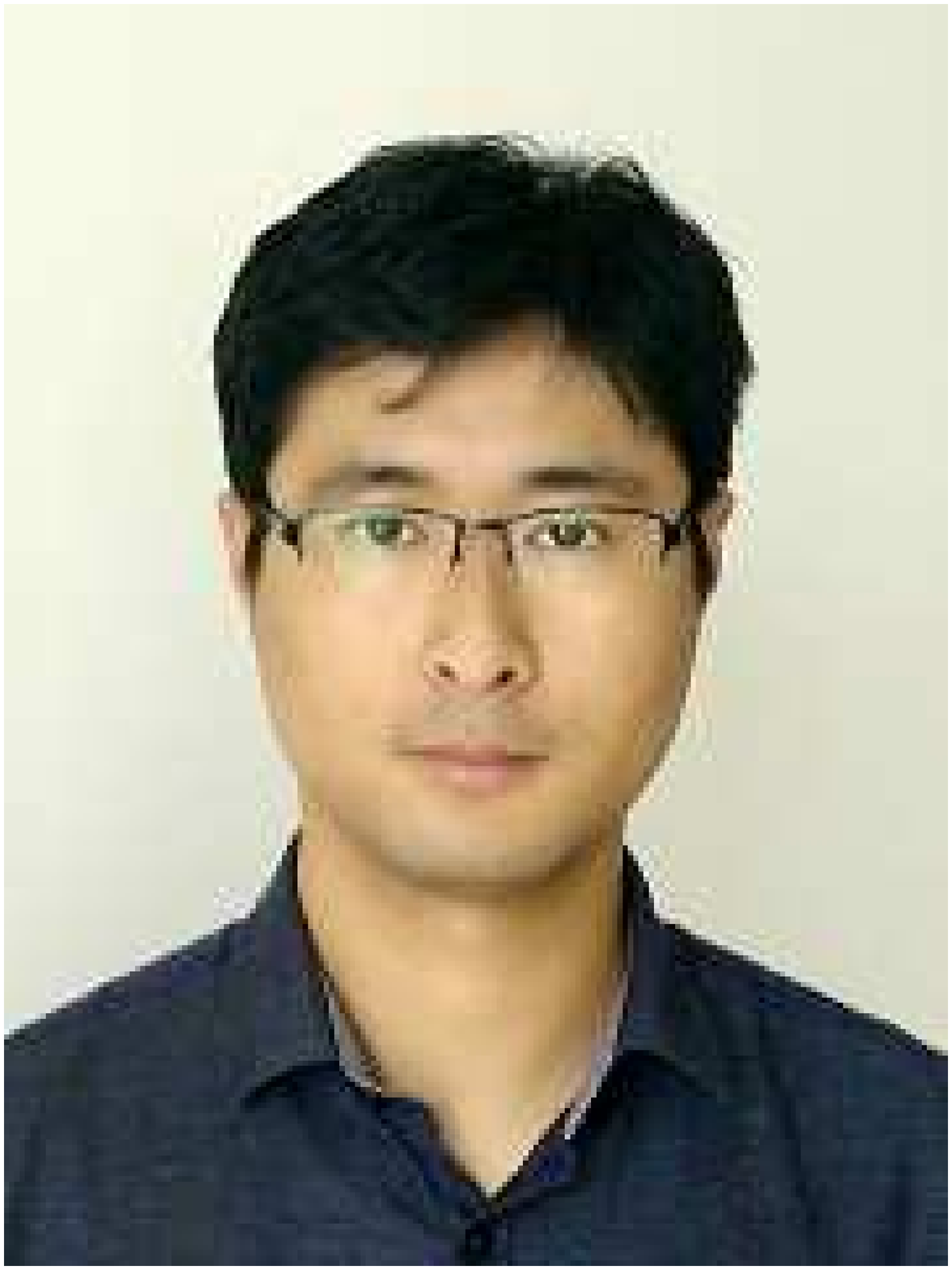}}]{Jie Xu}
is an Assistant Professor in the Department of Electrical and Computer Engineering at the University of Miami. He received his BS and MS degrees in Electronic Engineering from Tsinghua University in China in 2008 and 2010, respectively, and a PhD  degree in Electrical Engineering from University of California, Los Angeles (UCLA) in 2015. Dr. Xu's research interests are in game theory and learning theory with applications to education, communication, signal processing and network security. He received the Distinguished PhD Dissertation Award in Signals \& Systems at UCLA.
\end{IEEEbiography}

\begin{IEEEbiography}[{\includegraphics[width=1in,height=1.25in,clip,keepaspectratio]{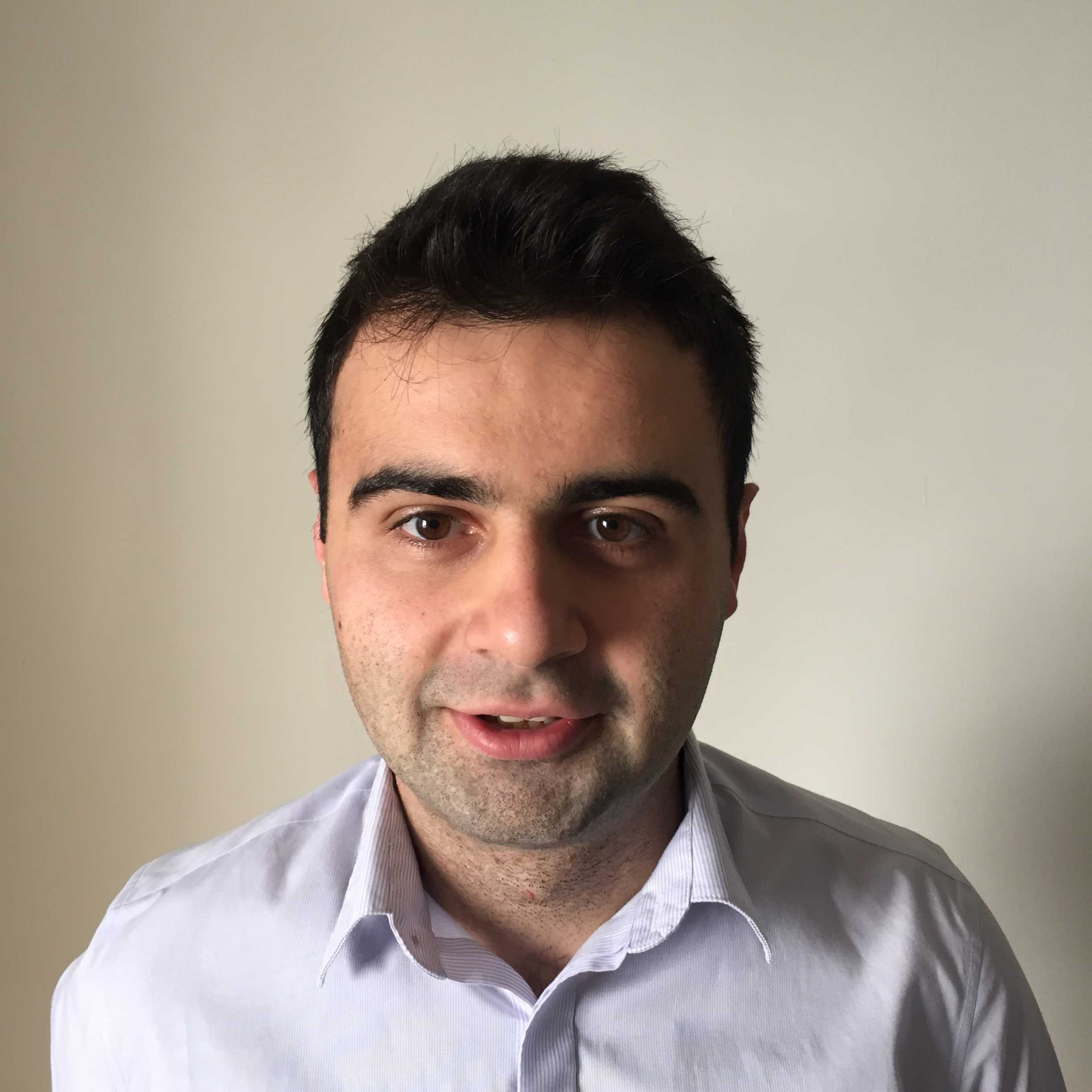}}]{Onur Atan}
received B.Sc. degree in Electrical Engineering from Bilkent University, Ankara, Turkey in 2013 and M.Sc. degree in Electrical Engineering from University of California, Los Angeles in 2014. He is currently pursuing the Ph.D. degree in Electrical Engineering at University of California, Los Angeles. He received the best M.Sc. thesis award in Electrical Engineering at University of California, Los Angeles.  His research interests include online learning and multi-armed bandit problems and their applications to medical informatics and education.
\end{IEEEbiography}

\begin{IEEEbiography}[{\includegraphics[width=1in,height=1.25in,clip,keepaspectratio]{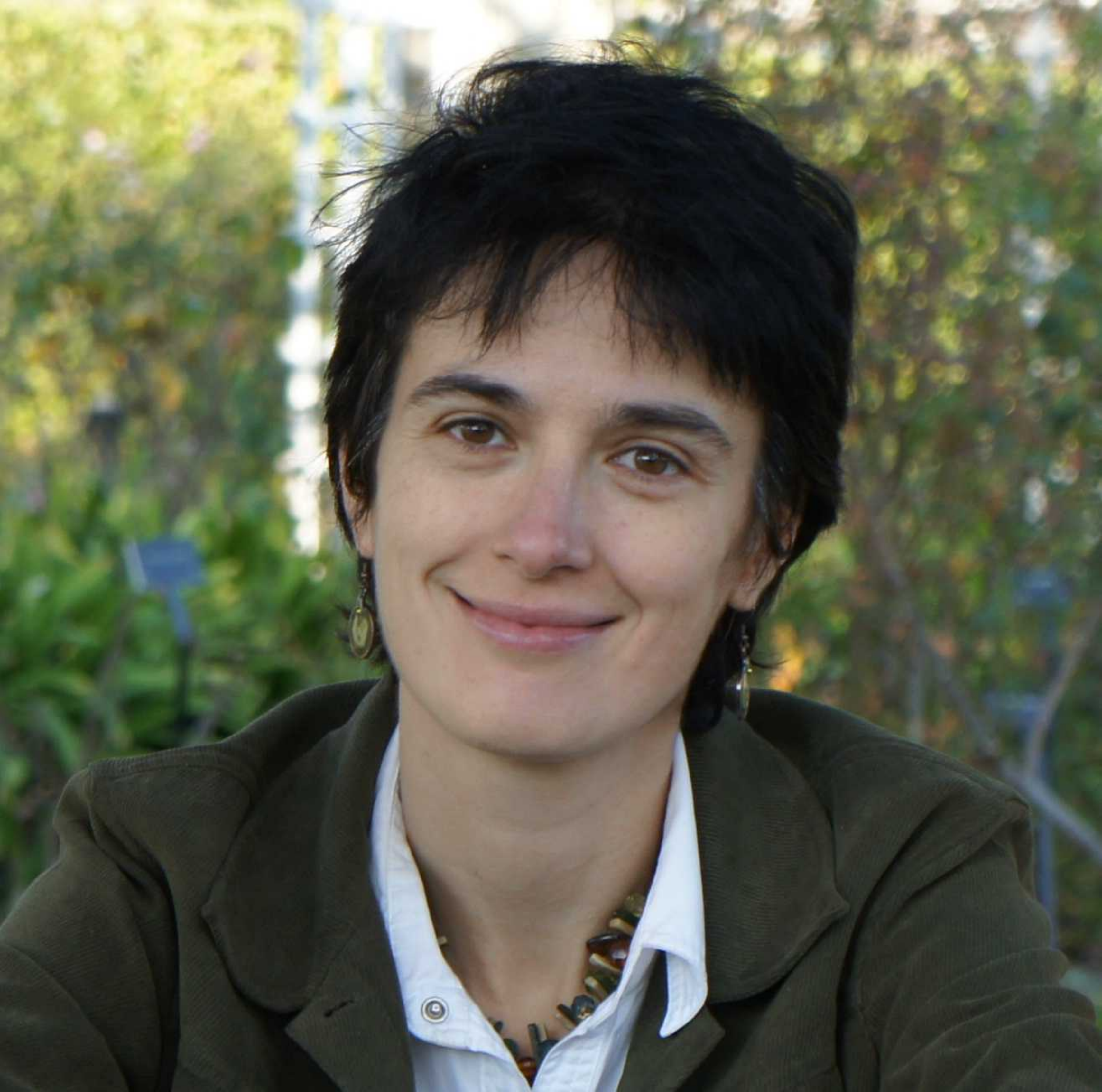}}]{Mihaela van der Schaar}
(F' 2009)  is Chancellor's Professor in the Electrical Engineering Department at UCLA. Her research interests include machine learning for medical informatics and education, online learning, stream mining, networks, network science, social networks and game theory. She received numerous awards, including the NSF Career Award, 3 IBM Faculty Awards, several best paper awards including the Darlington Best Paper Award. She has also 33 US patents.
\end{IEEEbiography}

%
%% if you will not have a photo at all:
%\begin{IEEEbiographynophoto}{John Doe}
%Biography text here.
%\end{IEEEbiographynophoto}

% insert where needed to balance the two columns on the last page with
% biographies
%\newpage

%\begin{IEEEbiographynophoto}{Jane Doe}
%Biography text here.
%\end{IEEEbiographynophoto}

% You can push biographies down or up by placing
% a \vfill before or after them. The appropriate
% use of \vfill depends on what kind of text is
% on the last page and whether or not the columns
% are being equalized.

%\vfill

% Can be used to pull up biographies so that the bottom of the last one
% is flush with the other column.
%\enlargethispage{-5in}

% that's all folks
\end{document}